%% file: main.tex
\documentclass[10pt]{article} 
\usepackage[preprint]{tmlr}

\usepackage{amsmath}
\usepackage{amsfonts}
\usepackage{amssymb}
\input{math_commands.tex}

\usepackage{hyperref}
\usepackage{url}
\usepackage{array}
\usepackage{booktabs}
\usepackage{multirow}
\usepackage{makecell}
\usepackage[table]{xcolor}
\usepackage{tikz}
\usetikzlibrary{arrows.meta, positioning, shapes.geometric, fit, backgrounds, mindmap}
\usepackage[edges]{forest}
\definecolor{hidden-draw}{RGB}{0,0,0}
\usepackage{graphicx}
\usepackage{enumitem}
\usepackage{diagbox}
\usepackage{xspace}
\usepackage{adjustbox}
\usepackage{fontawesome5}
\usepackage{tgheros}            
\usepackage[most]{tcolorbox}    
\input{figures/_palette.tex}

\definecolor{titledark}{HTML}{0F172A}  
\definecolor{titlegray}{HTML}{475569}  
\definecolor{titleline}{HTML}{CBD5E1}  
\definecolor{titlecard}{HTML}{F1F5F9}  
\definecolor{titleblue}{HTML}{2563EB}  
\newcommand{\titlesans}{\fontfamily{qhv}\selectfont}
\newcommand{\metalabel}[1]{{\titlesans\small\bfseries\color{titleblue}#1:}\hspace{0.35em}}

\newcommand{\ghlink}[1]{\href{#1}{\faGithub}}

\newcommand{\eg}{e.g.\xspace}

\newcommand{\worldmodel}{\mathcal{M}}

\newenvironment{takeaway}{%
  \par\smallskip
  \noindent\textbf{Takeaway.}\ \ignorespaces
}{%
  \par\smallskip
}

\def\month{MM}
\def\year{YYYY}
\def\openreview{\url{https://openreview.net/forum?id=XXXX}}

\newtcolorbox{abstractbox}{
  enhanced,
  frame hidden,
  colback=titlecard,
  colframe=titleline,
  boxrule=0.5pt,
  arc=8pt,
  left=0.6cm, right=0.6cm, top=0.5cm, bottom=0.25cm,
  before skip=0pt, after skip=1.2em,
  drop shadow={opacity=0.06}
}

\begin{document}

\thispagestyle{empty}

\vspace*{-1.1cm}

\vspace{1.0em}
\begin{center}
{\titlesans\bfseries\fontsize{21}{25}\selectfont\color{titledark}
Bridging the Agent-World Gap:\\
Text World Models for LLM-based Agents\par}
\vspace{1.1em}

{\normalsize\rmfamily\color{titledark}
Yixia Li$^{1}$ \hspace{0.6em}
Hongru Wang$^{2,*}$ \hspace{0.6em}
Peng Lai$^{1,*}$ \hspace{0.6em}
Zhiwen Ruan$^{1,*}$ \hspace{0.6em}
He Zhu$^{3,*}$ \hspace{0.6em}
Youxin Zhu$^{1}$\\[0.25em]
Ganlong Zhao$^{5}$ \hspace{0.6em}
Minda Hu$^{5}$ \hspace{0.6em}
Yun Chen$^{6}$ \hspace{0.6em}
Sibei Yang$^{4}$ \hspace{0.6em}
Peng Li$^{7}$ \hspace{0.6em}
Jeff Z.~Pan$^{2}$ \hspace{0.6em}
Jia Pan$^{8}$\\[0.25em]
Guanhua Chen$^{1,\dagger}$ \hspace{0.6em}
Yang Liu$^{7}$ \hspace{0.6em}
Guanbin Li$^{4,\dagger}$\par}
\vspace{0.22cm}

{\footnotesize\rmfamily\color{titlegray}
$^{1}$Southern University of Science and Technology \quad
$^{2}$University of Edinburgh \quad
$^{3}$Peking University\\[0.3em]
$^{4}$Sun Yat-sen University \quad
$^{5}$The Chinese University of Hong Kong \quad
$^{6}$Shanghai University of Finance and Economics\\[0.3em]
$^{7}$Tsinghua University \quad
$^{8}$The University of Hong Kong\par}
\end{center}

\vspace{0.45em}
\begin{abstractbox}
\setlength{\parindent}{0cm}
\raggedright
{\small\linespread{1.25}\selectfont\color{titledark}
\input{sections/00-abstract}\par}
\vspace{0.35em}
{\setlength{\parskip}{0.2em}\small
{\metalabel{Code}\href{https://github.com/sustech-nlp/awesome-text-world-models}{https://github.com/sustech-nlp/awesome-text-world-models}\par}
{\metalabel{Correspondence}%
\href{mailto:chengh3@sustech.edu.cn}{chengh3@sustech.edu.cn},
\href{mailto:liguanbin@mail.sysu.edu.cn}{liguanbin@mail.sysu.edu.cn}\par}
}
\vspace{0.25em}
{\footnotesize\rmfamily\itshape\color{titlegray}
$^{*}$Significant contributors. \\$^{\dagger}$Corresponding authors.\par}
\vspace{-1.5em}
\raggedleft
\includegraphics[height=1cm]{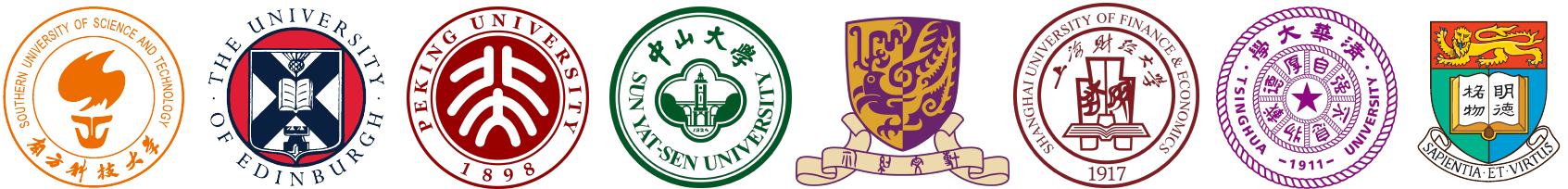}
\vspace{-0.5em}
\hspace*{-0.4cm}
\end{abstractbox}


\input{sections/01-introduction}

\input{sections/02-foundations}

\input{sections/03-building}

\input{sections/04-training-time}

\input{sections/05-inference-time}

\input{sections/06-evaluation}

\input{sections/07-open-problems}

\input{sections/08-conclusion}

\bibliography{references_venue}
\bibliographystyle{tmlr}



\end{document}

%% file: math_commands.tex
\usepackage{amsmath,amsfonts,bm}

\def\eqref#1{equation~\ref{#1}}

\def\1{\bm{1}}

\DeclareMathAlphabet{\mathsfit}{\encodingdefault}{\sfdefault}{m}{sl}
\SetMathAlphabet{\mathsfit}{bold}{\encodingdefault}{\sfdefault}{bx}{n}



%% file: figures/_palette.tex
\definecolor{morFound}{RGB}{160,115,180} 
\definecolor{morBuild}{RGB}{ 80,125,190} 
\definecolor{morTrain}{RGB}{210,130, 75} 
\definecolor{morInfer}{RGB}{ 95,165, 95} 
\definecolor{morEval} {RGB}{195, 85, 80} 
\definecolor{morAgent}{RGB}{ 90,140,165} 
\definecolor{morEnv}  {RGB}{105,150,105} 
\definecolor{morInk}  {RGB}{ 50, 50, 60} 
\definecolor{morGray} {RGB}{130,130,140} 

%% file: sections/00-abstract.tex
Large language model (LLM)-based agents are increasingly used in interactive textual environments, from web navigation and code editing to tool use and long-horizon dialogue. Yet many remain largely reactive, mapping observations to actions without an explicit model of how these environments are structured and evolve. This motivates \emph{text world models} (TWMs): transition models over textual states that, given a state and a candidate action, predict the resulting webpage, terminal output, API response, or user reply, thereby supporting planning, efficient learning, and principled evaluation. We systematically review text world models for LLM-based agents, organized around a formal framework and the agent lifecycle: (1)~\emph{Foundations}, defining text world models and characterizing them by state representation and grounding domain; (2)~\emph{Construction}, taxonomizing LLM-as-WM and code-as-WM paradigms and reviewing methods for building them; (3)~\emph{Application}, examining how world models support agents at training time through experience synthesis and at inference time through planning, verification, and adaptation; and (4)~\emph{Evaluation}, covering both evaluation of the world model itself and its use as an evaluation environment for agents. We aim to consolidate this rapidly developing area, clarify its design space, and highlight open challenges for future research.

%% file: sections/01-introduction.tex
\section{Introduction}
\label{sec:introduction}

Large language models (LLMs) now power a generation of autonomous agents that browse the web, edit code repositories, orchestrate tools, and converse with users over long horizons. Yet despite their fluency, many such agents remain largely reactive: each step maps the current observation to the next action, without an explicit and inspectable account of the environment as a structured, evolving system. A web agent may choose a link without modeling how pages and tasks evolve across clicks; a coding agent may patch a file without simulating how source changes propagate to runtime behavior; a dialogue agent may answer without maintaining a stable model of the user's evolving goals. These limitations point to a missing ingredient: a \textbf{world model} (WM), which captures how an environment is structured and evolves under an agent's actions, thereby supporting planning, efficient learning, and evaluation beyond one-step decisions.

LLM-based agents often operate through textual interfaces: webpages, repositories, APIs, terminals, and human interlocutors. This textual operating regime motivates \textbf{text world models} (TWMs): transition models over textual state. Here, textual state is used broadly to include structured task states, observable textual traces, and natural-language descriptions of the environment. Given such a state and a candidate action, a text world model predicts the resulting textual state, whether it is a webpage observation, a terminal output, an API response, or a user reply. This framing inherits the role world models have long played in reinforcement learning and control~\citep{ha2018worldmodels,hafner2024dreamerv3}, but the textual setting introduces tensions absent in pixel- or state-vector worlds. States are open-vocabulary and ambiguous; dynamics are knowledge-laden, with plausible transitions hinging on background facts about the text world; and correctness is subjective and semantic, rarely admitting a single right successor. Together, these tensions raise new questions for how text world models should be constructed, used, and evaluated.

\begin{figure}[t]
    \centering
    \includegraphics[width=0.90\textwidth]{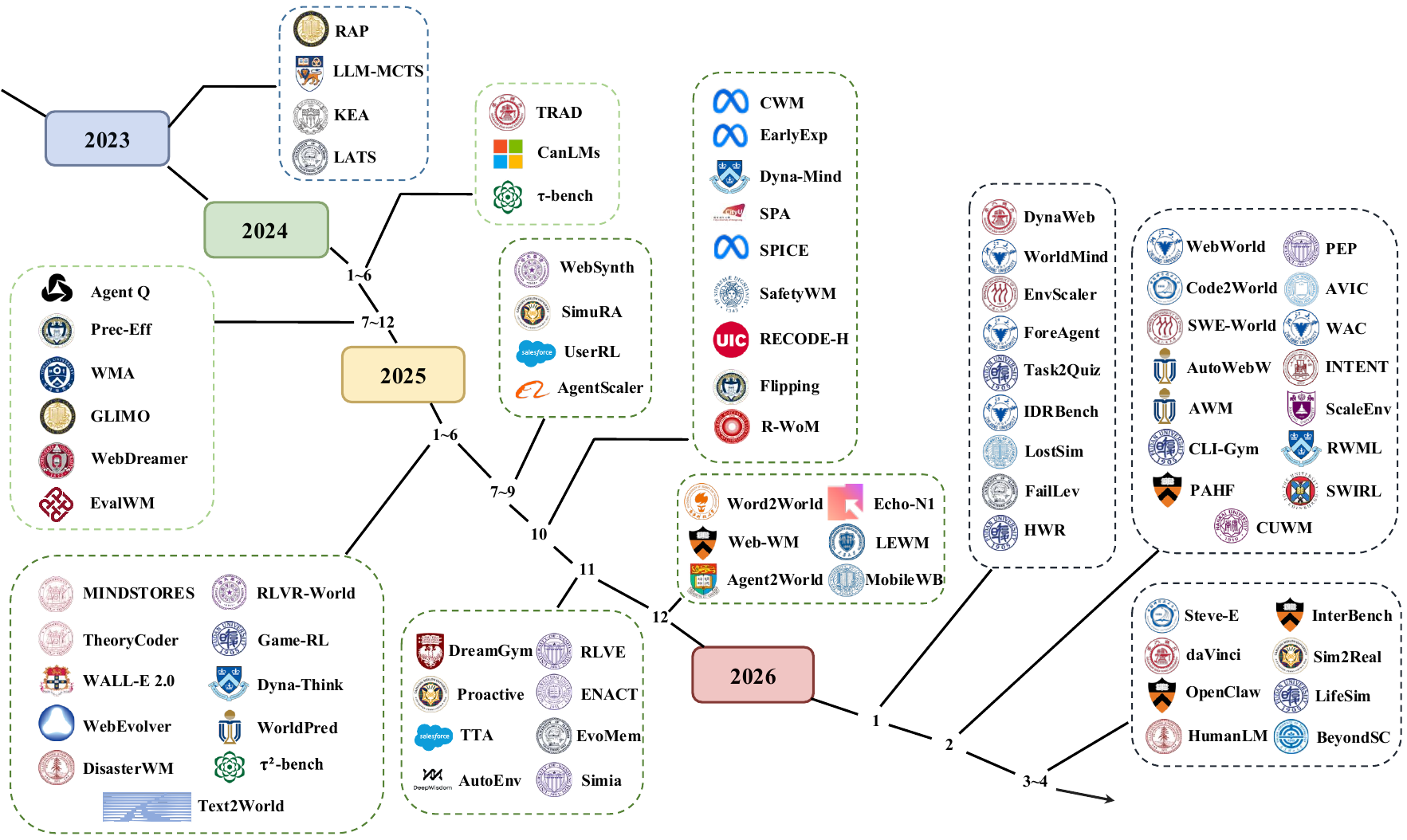}
    \caption{Chronological development of text world models for LLM-based agents. Each cluster groups representative works by release date; the field has expanded rapidly from 2025 onward.}
    \label{fig:timeline}
    \vspace{-15pt}
\end{figure}

Recent work has begun to answer these questions from multiple directions (Figure~\ref{fig:timeline}). On the construction side, text world models have been instantiated through prompt-based simulation~\citep{hao2023rap, gu2025webdreamer}, learned neural transition models~\citep{chae2025wma, xiao2025webworld}, and programmatic or hybrid constructions. On the use side, they serve as training-time data generators and environment substrates, as well as inference-time planners, verifiers, and rollout engines. Evaluation work has likewise started to move beyond one-step prediction toward consistency, behavior preservation, and downstream agent utility.

However, this emerging literature remains fragmented across web, code, tool-use, and dialogue domains, with little shared vocabulary and no systematic review of text world models as a distinct object of study. Existing world-model surveys predominantly target visual and control regimes,
including generative video and physics cognition~\citep{zhu2025soraworldsimulator},
autonomous driving~\citep{feng2025surveyworldmodelsautonomous},
and embodied intelligence~\citep{fung2025embodiedaiagentsmodeling,li2025comprehensivesurveyworldmodels}, while paying only passing attention to the text regime. Their classification axes and evaluation protocols, tuned for pixel- or state-vector worlds, do not transfer cleanly to textual transitions, whose ambiguity, knowledge dependence, and semantic fidelity requirements call for a dedicated taxonomy.

To address this gap, this survey provides the first systematic review of text world models for LLM-based agents. We focus on the lifecycle of a text world model in an agent stack: construction, usage, and evaluation. This lifecycle view provides a common framework for comparing methods that differ in how they are built, where they are deployed, and how they are evaluated. The survey contributes:
\begin{itemize}[leftmargin=1.2em,itemsep=2pt,topsep=2pt]
    \item \textbf{Foundations} (\S\ref{sec:foundations}): a formal framework for text world models and a two-axis characterization along state representation and grounding domain.
    \item \textbf{Construction} (\S\ref{sec:building}): a taxonomy of how text world models are built, covering learning-based, prompt-based, and programmatic approaches, together with a cross-paradigm comparison of when each is preferred.
    \item \textbf{Usage} (\S\ref{sec:training-time},~\S\ref{sec:inference-time}): a unified account of deployment, both at training time as substrates for agent learning and at inference time as planners, verifiers, and rollout engines.
    \item \textbf{Evaluation} (\S\ref{sec:evaluation}): an organization of evaluation methodology along single-step accuracy, multi-step consistency, behavior preservation, and downstream agent utility.
\end{itemize}
Figure~\ref{fig:mindmap} provides a compact map of the resulting taxonomy, linking the lifecycle stages to representative methods and papers surveyed in later sections.
Figure~\ref{fig:lifecycle} previews this structure, highlighting how construction choices, deployment roles, and evaluation criteria interact across the lifecycle. \S\ref{sec:open-problems} then synthesizes the limitations and open problems that cut across these stages.

\input{figures/lifecycle}
\input{figures/mindmap}

%% file: figures/lifecycle.tex
\begin{figure}
    \centering
    \resizebox{0.90\textwidth}{!}{%
    \begin{tikzpicture}[
        >=Stealth,
        font=\sffamily,
        numcircle/.style={
            circle, draw=#1, fill=#1, text=white,
            font=\sffamily\bfseries\footnotesize,
            inner sep=0pt, minimum size=14pt
        },
        stage/.style={
            rectangle, rounded corners=5pt,
            minimum width=2.6cm, minimum height=2.0cm,
            text width=2.4cm, align=center,
            font=\sffamily\small,
            draw=#1, fill=#1!12, line width=0.9pt,
            text=morInk
        },
        auxbox/.style={
            rectangle, rounded corners=4pt,
            minimum width=1.9cm, minimum height=0.9cm,
            text width=1.8cm, align=center,
            font=\sffamily\scriptsize,
            draw=#1, fill=#1!10, line width=0.7pt,
            text=morInk
        },
        arr/.style={->, line width=0.9pt, color=morGray},
        thickarr/.style={->, line width=1.2pt, color=#1},
        darr/.style={->, line width=0.8pt, dashed, color=#1!70},
        lbl/.style={font=\sffamily\fontsize{6}{7}\selectfont, text=morInk!70, fill=white, inner sep=1.5pt},
    ]

    \node[auxbox=morFound] (found) at (4.5, 2.1) {%
        \textcolor{morFound}{\faBookOpen}~\textbf{\S2 Foundations}\\[1pt]
        \fontsize{5.5}{6.5}\selectfont\textcolor{morInk!65}{Formalism, LLM-as-WM}};

    \node[stage=morBuild] (build) at (0, 0) {};
    \node[numcircle=morBuild, anchor=north west] at ([xshift=4pt, yshift=-4pt]build.north west) {1};
    \node[align=center, font=\sffamily\small\bfseries, text=morBuild]
        at ([yshift=8pt]build.north) {\S3 Build};
    \node[align=center, font=\sffamily\fontsize{6.5}{7.5}\selectfont, text=morInk!75]
        at (build.center) {\faCogs~Learning\\(SFT, RL)\\[1.5pt]
                            \faComments~Prompt\\(ICL, RAG)\\[1.5pt]
                            \faCode~Programmatic};

    \node[rectangle, rounded corners=7pt, draw=morInk!50, fill=white,
          minimum width=2.3cm, minimum height=1.3cm,
          font=\sffamily\small\bfseries, text=morInk, line width=1pt, align=center]
        (wm) at (4.5, 0) {\faGlobe~Text World\\Model $\mathcal{M}$};

    \node[stage=morTrain] (train) at (9, 1.4) {};
    \node[numcircle=morTrain, anchor=north west] at ([xshift=4pt, yshift=-4pt]train.north west) {2};
    \node[align=center, font=\sffamily\small\bfseries, text=morTrain]
        at ([yshift=8pt]train.north) {\S4 Train-Time};
    \node[align=center, font=\sffamily\fontsize{6.5}{7.5}\selectfont, text=morInk!75]
        at (train.center) {\faBrain~Internalize\\into params\\[1.5pt]
                           \faGamepad~Training\\environment\\[1.5pt]
                           \faUserFriends~User sim};

    \node[stage=morInfer, minimum height=1.5cm] (deploy) at (9, -1.4) {};
    \node[numcircle=morInfer, anchor=north west] at ([xshift=4pt, yshift=-4pt]deploy.north west) {3};
    \node[align=center, font=\sffamily\small\bfseries, text=morInfer]
        at ([yshift=8pt]deploy.north) {\S5 Inference-Time};
    \node[align=center, font=\sffamily\fontsize{6.5}{7.5}\selectfont, text=morInk!75]
        at (deploy.center) {\faTree~Simulator\\(lookahead)\\[1.5pt]
                            \faCheckCircle~Verifier\\(accept/revise)};

    \node[stage=morEval] (eval) at (13.5, 0) {};
    \node[numcircle=morEval, anchor=north west] at ([xshift=4pt, yshift=-4pt]eval.north west) {4};
    \node[align=center, font=\sffamily\small\bfseries, text=morEval]
        at ([yshift=8pt]eval.north) {\S6 Evaluate};
    \node[align=center, font=\sffamily\fontsize{6.5}{7.5}\selectfont, text=morInk!75]
        at (eval.center) {\faBullseye~Prediction\\fidelity\\[1.5pt]
                          \faChartLine~Task-driven\\utility\\[1.5pt]
                          \faFlask~WM as eval\\environment};

    \node[auxbox=morAgent] (agent) at (4.5, -2.9)
        {\textcolor{morAgent}{\faRobot}~\textbf{Agent $\pi$}};

    \node[auxbox=morEnv] (env) at (13.5, -2.9)
        {\textcolor{morEnv}{\faGlobeAmericas}~\textbf{Real Env.}\\[0pt]
         \fontsize{5.5}{6.5}\selectfont\textcolor{morInk!65}{Web, Code, Game}};

    \draw[arr, color=morFound!70] (found) -- node[lbl, right] {formalize} (wm.north);
    \draw[thickarr=morBuild] (build) -- node[lbl, above] {construct} (wm);

    \coordinate (fork) at ($(wm.east) + (0.6, 0)$);
    \draw[line width=1.2pt, color=morInk!40, -] (wm.east) -- (fork);
    \draw[thickarr=morTrain] (fork) -- ++(0, 1.4) node[lbl, above right, pos=1] {rollouts} -- (train.west);
    \draw[thickarr=morInfer] (fork) -- ++(0, -1.4) node[lbl, below right, pos=1] {lookahead} -- (deploy.west);

    \draw[arr, color=morTrain!70] (train.east) -- node[lbl, above, sloped, pos=0.45] {agent perf.} (eval.north west);
    \draw[arr, color=morInfer!70] (deploy.east) -- node[lbl, below, sloped, pos=0.45] {task metrics} (eval.south west);

    \draw[arr, color=morTrain!50]
        ([yshift=-12pt]train.west) -- ++(-0.4, 0) |- (4.5, -1.0)
        -- node[lbl, right, pos=0.45] {train $\pi$} (agent.north);
    \draw[arr, color=morAgent!70]
        (agent) -| node[lbl, pos=0.25, above] {act} (deploy);
    \draw[arr, color=morEnv!70]
        (deploy) |- node[lbl, pos=0.7, above] {execute} (env);
    \draw[arr, color=morEnv!70]
        ([xshift=4pt]env.north) -- node[lbl, right] {feedback} ([xshift=4pt]eval.south);

    \draw[darr=morEval]
        ([xshift=18pt]eval.south) -- ++(0, -2.5) -| node[lbl, pos=0.25, above] {\textcolor{morEval!80}{improve WM}} (build.south);

    \node[auxbox=morEnv] at (13.5, -2.9)
        {\textcolor{morEnv}{\faGlobeAmericas}~\textbf{Real Env.}\\[0pt]
         \fontsize{5.5}{6.5}\selectfont\textcolor{morInk!65}{Web, Code, Game}};

    \end{tikzpicture}}
    \caption{The text world model lifecycle. A world model $\mathcal{M}$ is first \emph{constructed} via learning, prompting, or code generation (\S\ref{sec:building}); then used to \emph{train} agents through synthetic rollouts (\S\ref{sec:training-time}) and \emph{guide} them via lookahead at inference time (\S\ref{sec:inference-time}); finally \emph{evaluated} for fidelity and utility (\S\ref{sec:evaluation}). The dashed arrow indicates a feedback loop that iteratively improves the world model.}
    \label{fig:lifecycle}
\end{figure}
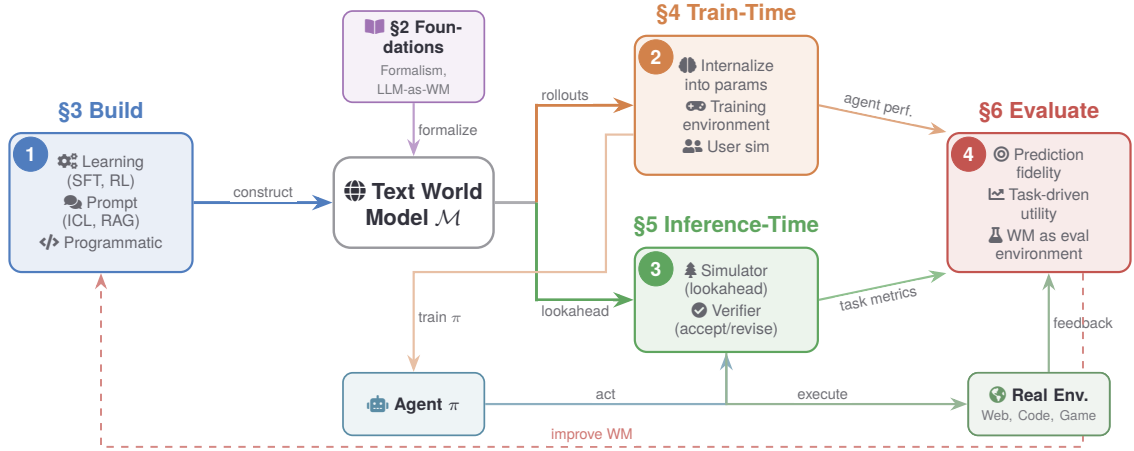

%% file: figures/mindmap.tex
\definecolor{mmFound}{RGB}{180,130,200} 
\definecolor{mmBuild}{RGB} { 90,140,210} 
\definecolor{mmTrain}{RGB}{230,140, 75} 
\definecolor{mmInfer}{RGB}{105,180,105} 
\definecolor{mmEval} {RGB}{215, 90, 85} 
\definecolor{mmRoot} {RGB}{120,120,135} 
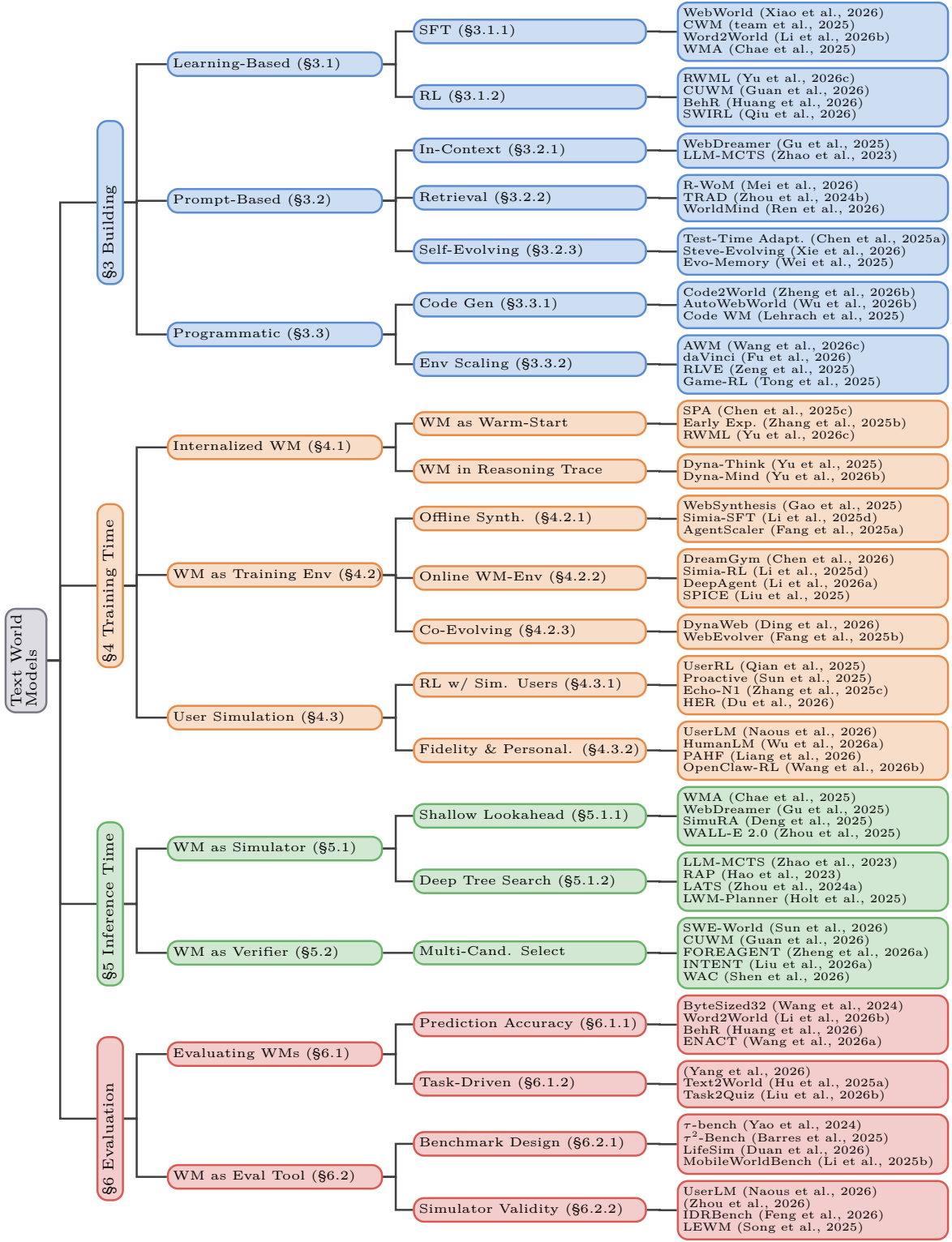
\begin{figure}[p]
    \centering
    \vspace*{-.5cm}
    \tikzstyle{my-box}=[
    rectangle,
    rounded corners,
    text opacity=1,
    minimum height=0.1em,
    minimum width=0.1em,
    inner sep=2pt,
    align=left,
    fill opacity=.5,
    ]
    \tikzstyle{leaf}=[my-box]
    \resizebox{0.95\textwidth}{!}{
        \begin{forest}
            forked edges,
            for tree={
                grow=east,
                reversed=true,
                anchor=base west,
                parent anchor=east,
                child anchor=west,
                base=left,
                font=\fontsize{6}{6}\selectfont,
                rectangle,
                draw=hidden-draw,
                rounded corners,
                align=left,
                minimum width=0.1em,
                edge+={darkgray, line width=0.8pt},
                s sep=2pt,
                inner xsep=2pt,
                inner ysep=2pt,
                ver/.style={rotate=90, child anchor=north, parent anchor=south, anchor=center},
            },
            where level=1{text width=6em,font=\fontsize{6}{6}\selectfont,align=center,rotate=90, anchor=north, child anchor=north, parent anchor=south}{},
            where level=2{text width=8em,font=\fontsize{5}{5}\selectfont}{},
            where level=3{text width=8.7em,font=\fontsize{5}{5}\selectfont,}{},
            where level=4{text width=10.2em,font=\fontsize{4.5}{4.5}\selectfont,}{},
            [
                Text World \\ Models, draw=hidden-draw, color=mmRoot, fill=mmRoot!25, thick, text=black, ver
                [
                    \S3~Building
                    , color=mmBuild, fill=mmBuild!30, thick, text=black
                    [
                        Learning-Based (\S\ref{sec:learning-based})
                        , color=mmBuild, fill=mmBuild!30, thick, text=black
                        [
                            SFT (\S\ref{sec:sft-wm})
                            , color=mmBuild, fill=mmBuild!30, thick, text=black
                            [
                                WebWorld~\citep{xiao2025webworld}\\
                                CWM~\citep{faircwm2025}\\
                                Word2World~\citep{li2025wordtoworld}\\
                                WMA~\citep{chae2025wma}
                                , color=mmBuild, fill=mmBuild!30, thick, text=black
                            ]
                        ]
                        [
                            RL (\S\ref{sec:rl-wm})
                            , color=mmBuild, fill=mmBuild!30, thick, text=black
                            [
                                RWML~\citep{yu2025rwml}\\
                                CUWM~\citep{guan2026computerusingworldmodel}\\
                                BehR~\citep{huang2026stateconsistencybehaviorconsistency}\\
                                SWIRL~\citep{qiu2026selfimprovingworldmodellinglatent}
                                , color=mmBuild, fill=mmBuild!30, thick, text=black
                            ]
                        ]
                    ]
                    [
                        Prompt-Based (\S\ref{sec:prompt-based})
                        , color=mmBuild, fill=mmBuild!30, thick, text=black
                        [
                            In-Context (\S\ref{sec:incontext-wm})
                            , color=mmBuild, fill=mmBuild!30, thick, text=black
                            [
                                WebDreamer~\citep{gu2025webdreamer}\\
                                LLM-MCTS~\citep{zhao2023llmmcts}
                                , color=mmBuild, fill=mmBuild!30, thick, text=black
                            ]
                        ]
                        [
                            Retrieval (\S\ref{sec:retrieval-wm})
                            , color=mmBuild, fill=mmBuild!30, thick, text=black
                            [
                                R-WoM~\citep{mei2025rwom}\\
                                TRAD~\citep{zhou2024trad}\\
                                WorldMind~\citep{ren2025worldmind}
                                , color=mmBuild, fill=mmBuild!30, thick, text=black
                            ]
                        ]
                        [
                            Self-Evolving (\S\ref{sec:self-evolving-wm})
                            , color=mmBuild, fill=mmBuild!30, thick, text=black
                            [
                                Test-Time Adapt.~\citep{chen2025testtimeadapt}\\
                                Steve-Evolving~\citep{xie2025steveevolving}\\
                                Evo-Memory~\citep{wei2025evomemory}
                                , color=mmBuild, fill=mmBuild!30, thick, text=black
                            ]
                        ]
                    ]
                    [
                        Programmatic (\S\ref{sec:programmatic})
                        , color=mmBuild, fill=mmBuild!30, thick, text=black
                        [
                            Code Gen (\S\ref{sec:generated-envs})
                            , color=mmBuild, fill=mmBuild!30, thick, text=black
                            [
                                Code2World~\citep{zheng2025code2world}\\
                                AutoWebWorld~\citep{wu2025autowebworld}\\
                                Code WM~\citep{lehrach2025codewm}
                                , color=mmBuild, fill=mmBuild!30, thick, text=black
                            ]
                        ]
                        [
                            Env Scaling (\S\ref{sec:env-scaling})
                            , color=mmBuild, fill=mmBuild!30, thick, text=black
                            [
                                AWM~\citep{wang2025awm}\\
                                daVinci~\citep{fu2025davincienv}\\
                                RLVE~\citep{zeng2025rlve}\\
                                Game-RL~\citep{tong2025gamerl}
                                , color=mmBuild, fill=mmBuild!30, thick, text=black
                            ]
                        ]
                    ]
                ]
                [
                    \S4~Training Time
                    , color=mmTrain, fill=mmTrain!30, thick, text=black
                    [
                        Internalized WM (\S\ref{subsec:internalized-wm})
                        , color=mmTrain, fill=mmTrain!30, thick, text=black
                        [
                            WM as Warm-Start
                            , color=mmTrain, fill=mmTrain!30, thick, text=black
                            [
                                SPA~\citep{chen2025spa}\\
                                Early~Exp.~\citep{zhang2025earlyexperience}\\
                                RWML~\citep{yu2025rwml}
                                , color=mmTrain, fill=mmTrain!30, thick, text=black
                            ]
                        ]
                        [
                            WM in Reasoning Trace
                            , color=mmTrain, fill=mmTrain!30, thick, text=black
                            [
                                Dyna-Think~\citep{yu2025dynathink}\\
                                Dyna-Mind~\citep{yu2025dynamind}
                                , color=mmTrain, fill=mmTrain!30, thick, text=black
                            ]
                        ]
                    ]
                    [
                        WM as Training Env (\S\ref{subsec:llm-env-substrate})
                        , color=mmTrain, fill=mmTrain!30, thick, text=black
                        [
                            Offline Synth. (\S\ref{subsubsec:offline-synth})
                            , color=mmTrain, fill=mmTrain!30, thick, text=black
                            [
                                WebSynthesis~\citep{gao2025websynthesis}\\
                                Simia-SFT~\citep{li2025simia}\\
                                AgentScaler~\citep{fang2025agentscaler}
                                , color=mmTrain, fill=mmTrain!30, thick, text=black
                            ]
                        ]
                        [
                            Online WM-Env (\S\ref{subsubsec:online-llm-env})
                            , color=mmTrain, fill=mmTrain!30, thick, text=black
                            [
                                DreamGym~\citep{chen2025dreamgym}\\
                                Simia-RL~\citep{li2025simia}\\
                                DeepAgent~\citep{li2026deepagent}\\
                                SPICE~\citep{liu2025spice}
                                , color=mmTrain, fill=mmTrain!30, thick, text=black
                            ]
                        ]
                        [
                            Co-Evolving (\S\ref{subsubsec:coevol-wm})
                            , color=mmTrain, fill=mmTrain!30, thick, text=black
                            [
                                DynaWeb~\citep{ding2026dynaweb}\\
                                WebEvolver~\citep{fang2025webevolver}
                                , color=mmTrain, fill=mmTrain!30, thick, text=black
                            ]
                        ]
                    ]
                    [
                        User Simulation (\S\ref{subsec:user-sim-training})
                        , color=mmTrain, fill=mmTrain!30, thick, text=black
                        [
                            RL w/ Sim. Users (\S\ref{subsubsec:user-rl})
                            , color=mmTrain, fill=mmTrain!30, thick, text=black
                            [
                                UserRL~\citep{qian2025userrl}\\
                                Proactive~\citep{sun2025proactive}\\
                                Echo-N1~\citep{zhang2025echon1}\\
                                HER~\citep{du2026her}
                                , color=mmTrain, fill=mmTrain!30, thick, text=black
                            ]
                        ]
                        [
                            Fidelity \& Personal. (\S\ref{subsubsec:personalization})
                            , color=mmTrain, fill=mmTrain!30, thick, text=black
                            [
                                UserLM~\citep{naous2025flipping}\\
                                HumanLM~\citep{wu2026humanlmsimulatingusersstate}\\
                                PAHF~\citep{liang2026pahf}\\
                                OpenClaw-RL~\citep{wang2026openclawrl}
                                , color=mmTrain, fill=mmTrain!30, thick, text=black
                            ]
                        ]
                    ]
                ]
                [
                    \S5~Inference Time
                    , color=mmInfer, fill=mmInfer!30, thick, text=black
                    [
                        WM as Simulator (\S\ref{subsec:wm-as-simulator})
                        , color=mmInfer, fill=mmInfer!30, thick, text=black
                        [
                            Shallow Lookahead (\S\ref{subsubsec:shallow-lookahead})
                            , color=mmInfer, fill=mmInfer!30, thick, text=black
                            [
                                WMA~\citep{chae2025wma}\\
                                WebDreamer~\citep{gu2025webdreamer}\\
                                SimuRA~\citep{deng2025simuraworldmodeldrivensimulativereasoning}\\
                                WALL-E~2.0~\citep{zhou2025walle2}
                                , color=mmInfer, fill=mmInfer!30, thick, text=black
                            ]
                        ]
                        [
                            Deep Tree Search (\S\ref{subsubsec:tree-search})
                            , color=mmInfer, fill=mmInfer!30, thick, text=black
                            [
                                LLM-MCTS~\citep{zhao2023llmmcts}\\
                                RAP~\citep{hao2023rap}\\
                                LATS~\citep{zhou2024lats}\\
                                LWM-Planner~\citep{holt2025improvingllmagentplanning}
                                , color=mmInfer, fill=mmInfer!30, thick, text=black
                            ]
                        ]
                    ]
                    [
                        WM as Verifier (\S\ref{subsec:wm-as-verifier})
                        , color=mmInfer, fill=mmInfer!30, thick, text=black
                        [
                            Multi-Cand. Select
                            , color=mmInfer, fill=mmInfer!30, thick, text=black
                            [
                                SWE-World~\citep{sun2025sweworld}\\
                                CUWM~\citep{guan2026computerusingworldmodel}\\
                                FOREAGENT~\citep{zheng2026foreagent}\\
                                INTENT~\citep{liu2026intent}\\
                                WAC~\citep{shen2026wac}
                                , color=mmInfer, fill=mmInfer!30, thick, text=black
                            ]
                        ]
                    ]
                ]
                [
                    \S6~Evaluation
                    , color=mmEval, fill=mmEval!30, thick, text=black
                    [
                        Evaluating WMs (\S\ref{subsec:eval-wm})
                        , color=mmEval, fill=mmEval!30, thick, text=black
                        [
                            Prediction Accuracy (\S\ref{subsubsec:prediction-accuracy})
                            , color=mmEval, fill=mmEval!30, thick, text=black
                            [
                                ByteSized32~\citep{wang2024canlms}\\
                                Word2World~\citep{li2025wordtoworld}\\
                                BehR~\citep{huang2026stateconsistencybehaviorconsistency}\\
                                ENACT~\citep{wang2025enactevaluatingembodiedcognition}
                                , color=mmEval, fill=mmEval!30, thick, text=black
                            ]
                        ]
                        [
                            Task-Driven (\S\ref{subsubsec:task-driven-eval})
                            , color=mmEval, fill=mmEval!30, thick, text=black
                            [
                                \citep{yang2026evalwm}\\
                                Text2World~\citep{hu2025text2world}\\
                                Task2Quiz~\citep{liu2026task2quiz}
                                , color=mmEval, fill=mmEval!30, thick, text=black
                            ]
                        ]
                    ]
                    [
                        WM as Eval Tool (\S\ref{subsec:wm-as-eval})
                        , color=mmEval, fill=mmEval!30, thick, text=black
                        [
                            Benchmark Design (\S\ref{subsubsec:benchmark-design})
                            , color=mmEval, fill=mmEval!30, thick, text=black
                            [
                                $\tau$-bench~\citep{yao2024taubenchbenchmarktoolagentuserinteraction}\\
                                $\tau^2$-Bench~\citep{barres2025tau2bench}\\
                                LifeSim~\citep{duan2026lifesimlonghorizonuserlife}\\
                                MobileWorldBench~\citep{li2025mobileworldbench}
                                , color=mmEval, fill=mmEval!30, thick, text=black
                            ]
                        ]
                        [
                            Simulator Validity (\S\ref{subsubsec:simulator-validity})
                            , color=mmEval, fill=mmEval!30, thick, text=black
                            [
                                UserLM~\citep{naous2025flipping}\\
                                \citep{zhou2026mindsim2realgapuser}\\
                                IDRBench~\citep{feng2026idrbench}\\
                                LEWM~\citep{song2025largeemotionalworldmodel}
                                , color=mmEval, fill=mmEval!30, thick, text=black
                            ]
                        ]
                    ]
                ]
            ]
        \end{forest}
        }
    \caption{Taxonomy of text world model research organized by the agent lifecycle. Colors encode lifecycle stages: \textcolor{mmBuild}{\textbf{blue}} = construction (\S\ref{sec:building}), \textcolor{mmTrain}{\textbf{orange}} = training-time use (\S\ref{sec:training-time}), \textcolor{mmInfer}{\textbf{green}} = inference-time deployment (\S\ref{sec:inference-time}), \textcolor{mmEval}{\textbf{red}} = evaluation (\S\ref{sec:evaluation}). Representative papers are shown at the leaves.}
    \label{fig:mindmap}
\end{figure}

%% file: sections/02-foundations.tex
\section{Foundations and Formalism}
\label{sec:foundations}

This section fixes the conceptual object studied in the rest of the survey. The lifecycle in Figure~\ref{fig:lifecycle} organizes what researchers do with text world models---constructing, deploying, and evaluating them. Here we instead ask what kind of object a text world model is. We first define it as a textual transition model and delimit the literature covered (\S\ref{sec:twm-definition}), and then introduce an object-level taxonomy that cuts across the later lifecycle stages (\S\ref{sec:twm-axes}).
\subsection{Definition and Scope}
\label{sec:twm-definition}

We define a \emph{text world model} at the level of the transition function. Given an agent-visible state $s_t$ and a candidate action $a_t$, it returns a textual rendering of the successor state:
\[
\begin{aligned}
\worldmodel &: \mathcal{S} \times \mathcal{A} \to \mathcal{T}_{\mathcal{S}},\\
\hat{s}_{t+1} &= \worldmodel(s_t,a_t),
\end{aligned}
\]
where $\mathcal{T}_{\mathcal{S}}$ denotes textual renderings of successor states. The output $\hat{s}_{t+1}$ may take any textual form---a natural-language description, a structured record, a code fragment, or any other artifact that specifies what holds after executing $a_t$ in $s_t$. The qualifier ``text'' thus refers to the \emph{form} in which the transition is exposed to the agent, whereas the input state itself may be textual or multimodal, such as a dialogue history, a terminal trace, or a webpage screenshot. The underlying substrate is likewise unconstrained: a text world model may be elicited from a frozen LLM via prompting, trained as a dedicated transition predictor, encoded as executable program logic, or realized through other means.

This transition-centric view sets text world models apart from both classical latent-space world models and ordinary language models. Unlike world models that predict pixels, embeddings, or compact symbolic states~\citep{ha2018worldmodels,schrittwieser2020mastering,hafner2024dreamerv3}, a text world model surfaces its simulated dynamics as text that can be read, edited, verified, and fed back into an LLM agent. Unlike a generic LLM response, its role is explicitly counterfactual: given a state--action pair, it predicts how the interaction state \emph{would} change if that action were taken.

This survey covers two interaction targets that an agent may face: the \emph{external environment} and the \emph{human user}. Environment modeling targets systems such as web interfaces, software repositories, or tool APIs, where the dynamics manifest as page updates, file changes, or state transitions of a deployed runtime. User modeling targets the evolution of human-side variables in interaction, including preferences, intentions, and task progress over the course of a dialogue.

\subsection{A Two-Axis Taxonomy}
\label{sec:twm-axes}

Building on the definition and scope above, we propose a two-dimensional taxonomy of text world models along the axes of \emph{state representation} and \emph{grounding domain}, and map existing works onto it in Figure~\ref{fig:paper-map}. The former characterizes the textual form in which a model expresses states and transitions, while the latter characterizes the type of world whose dynamics the model seeks to capture.

\input{figures/paper-map}

\paragraph{Axis 1: State/transition representation} Three categories form a spectrum from flexible to formal:
\begin{enumerate}[leftmargin=*, label=(\arabic*)]
    \item \textbf{Natural language} renders states and transitions as free-form descriptions, summaries, or rationales. It is easy for LLMs to produce and consume, but it weakly constrains consistency, completeness, and executability.
    \item \textbf{Structured representations} impose an explicit schema, e.g., JSON records, key--value stores, knowledge graphs, accessibility trees, or PDDL-style predicates, exposing entities, relations, preconditions, and effects in a form that is easier to track and verify.
    \item \textbf{Executable code} encodes part or all of the transition as Python, TypeScript, HTML, simulators, or planning operators. When the domain admits precise operational semantics, execution replaces free-form prediction and yields stronger reproducibility and constraint enforcement.
\end{enumerate}

\paragraph{Axis 2: Grounding domain} The ``world'' that a text world model simulates varies greatly across applications:
\begin{enumerate}[leftmargin=*, label=(\arabic*)]
    \item \textbf{Physical worlds} are governed by embodied or physical regularities, as in household environments, navigation, game physics, and disaster scenes, where valid successors depend on commonsense or simulated physics.
    \item \textbf{Digital worlds} are governed by deployed computational systems, including websites, operating systems, code repositories, terminals, tool APIs, and GUI applications.
    \item \textbf{Social worlds} are governed by human or human-like behavior, spanning utterances, preferences, affect, cooperation, and task progress in dialogue.
    \item \textbf{Abstract worlds} are governed by formal or symbolic rules specified independently of any deployed runtime, such as planning domains, mathematical environments, game-theoretic abstractions, and logic puzzles.
\end{enumerate}

%% file: figures/paper-map.tex
\begin{figure*}[t]
    \centering
    \resizebox{\textwidth}{!}{%
    \footnotesize
    \renewcommand{\arraystretch}{1.15}
    \setlength{\tabcolsep}{2.5pt}
    \begin{tabular}{@{}>{\centering\arraybackslash}p{1.6cm}|>{\raggedright\arraybackslash}p{3.5cm}|>{\raggedright\arraybackslash}p{3.5cm}|>{\raggedright\arraybackslash}p{3.5cm}|>{\raggedright\arraybackslash}p{3.5cm}@{}}
    \multicolumn{1}{c|}{\diagbox[width=1.5cm, height=1cm]{\tiny\textbf{State}}{\tiny\textbf{Domain}}}
    & \cellcolor{blue!8}\centering\textbf{\faCubes~Physical}
    & \cellcolor{orange!8}\centering\textbf{\faLaptop~Digital}
    & \cellcolor{green!8}\centering\textbf{\faUsers~Social}
    & \cellcolor{violet!8}\centering\textbf{\faBrain~Abstract} \tabularnewline
    \hline
    \cellcolor{red!8}\centering\textbf{\faCommentDots\\Natural Language}
    & \cellcolor{blue!4}\tiny
        Dyna-Mind~\citep{yu2025dynamind},
        LLM-MCTS~\citep{zhao2023llmmcts},
        MINDSTORES~\citep{chari2025mindstores},
        Disaster~WM~\citep{li2025disasterwm},
        Steve-Evolving~\citep{xie2025steveevolving},
        Word2World~\citep{li2025wordtoworld},
        WorldMind~\citep{ren2025worldmind},
        SWIRL~\citep{qiu2026selfimprovingworldmodellinglatent}
    & \cellcolor{orange!4}\tiny
        WebDreamer~\citep{gu2025webdreamer},
        WMA~\citep{chae2025wma},
        CUWM~\citep{guan2026computerusingworldmodel},
        Dyna-Think~\citep{yu2025dynathink},
        DreamGym~\citep{chen2025dreamgym},
        R-WoM~\citep{mei2025rwom},
        TRAD~\citep{zhou2024trad},
        WAC~\citep{shen2026wac},
        INTENT~\citep{liu2026intent},
        Simia~\citep{li2025simia},
        LATS~\citep{zhou2024lats},
        Evo-Memory~\citep{wei2025evomemory},
        SWIRL~\citep{qiu2026selfimprovingworldmodellinglatent}
    & \cellcolor{green!4}\tiny
        UserRL~\citep{qian2025userrl},
        Echo-N1~\citep{zhang2025echon1},
        HER~\citep{du2026her},
        OpenClaw-RL~\citep{wang2026openclawrl},
        PAHF~\citep{liang2026pahf},
        UserLM~\citep{naous2025flipping},
        RECODE-H~\citep{miao2025recodeh},
        IDRBench~\citep{feng2026idrbench}
    & \cellcolor{violet!4}\tiny
        RAP~\citep{hao2023rap},
        LATS~\citep{zhou2024lats},
        SPICE~\citep{liu2025spice},
        FOREAGENT~\citep{zheng2026foreagent},
        Evo-Memory~\citep{wei2025evomemory},
        Task2Quiz~\citep{liu2026task2quiz}
    \tabularnewline
    \hline
    \cellcolor{red!8}\centering\textbf{\faProjectDiagram\\Structured}
    & \cellcolor{blue!4}\tiny
        WorMI~\citep{yoo2025wormi},
        WALL-E~2.0~\citep{zhou2025walle2},
        ByteSized32~\citep{wang2024canlms},
        AEC~\citep{yang2025aec}
    & \cellcolor{orange!4}\tiny
        WebWorld~\citep{xiao2025webworld},
        WebEvolver~\citep{fang2025webevolver},
        DynaWeb~\citep{ding2026dynaweb},
        WebSynthesis~\citep{gao2025websynthesis},
        RLVR-World~\citep{wu2025rlvrworld},
        SWE-World~\citep{sun2025sweworld},
        CWM~\citep{faircwm2025},
        DeepAgent~\citep{li2026deepagent}
    & \cellcolor{green!4}\tiny
        $\tau^2$-Bench~\citep{barres2025tau2bench},
        Pep~\citep{bose2026pep},
        DWM~\citep{huang2024discretewm}
    & \cellcolor{violet!4}\tiny
        Text2World~\citep{hu2025text2world},
        SPA~\citep{chen2025spa}
    \tabularnewline
    \hline
    \cellcolor{red!8}\centering\textbf{\faCode\\Executable Code}
    & \cellcolor{blue!4}\tiny
        Game-RL~\citep{tong2025gamerl},
        TheoryCoder~\citep{ahmed2025theorycoder},
        WALL-E~2.0~\citep{zhou2025walle2},
        Agent2World~\citep{hu2025agent2world}
    & \cellcolor{orange!4}\tiny
        Code2World~\citep{zheng2025code2world},
        AutoWebWorld~\citep{wu2025autowebworld},
        CLI-Gym~\citep{lin2025cligym},
        AWM~\citep{wang2025awm},
        EnvScaler~\citep{song2025envscaler},
        ScaleEnv~\citep{tu2025scaleenv},
        daVinci~\citep{fu2025davincienv},
        AgentScaler~\citep{fang2025agentscaler},
        Web~WMs~\citep{feng2025webworldmodels},
        RLVE~\citep{zeng2025rlve},
        TheoryCoder~\citep{ahmed2025theorycoder}
    & \cellcolor{green!4}\tiny
        \mbox{}
    & \cellcolor{violet!4}\tiny
        Code~WM~\citep{lehrach2025codewm},
        Agent2World~\citep{hu2025agent2world},
        Game-RL~\citep{tong2025gamerl},
        AutoEnv~\citep{zhang2025autoenv}
    \tabularnewline
    \end{tabular}%
    }
    \caption{Two-dimensional landscape of text world models along state representation and grounding domain.
    The horizontal axis categorizes \textbf{grounding domain}---\emph{Physical} (embodied environments, game physics, disaster assessment), \emph{Digital} (web interfaces, software, code reasoning), \emph{Social} (user simulation, dialogue), and \emph{Abstract} (planning, math, and symbolic reasoning).
    The vertical axis categorizes \textbf{state representation}---\emph{Natural Language} (free-form text descriptions), \emph{Structured} (JSON, knowledge graphs, accessibility trees), and \emph{Executable Code} (Python, TypeScript, HTML programs).}
    \label{fig:paper-map}
\end{figure*}

%% file: sections/03-building.tex
\section{Building Text World Models}
\label{sec:building}

This section examines how text world models are built. We organize existing approaches by where the state transition is actually carried out, which yields two top-level paradigms. In the LLM-as-world-model paradigm, the transition function is the LLM's own forward pass, and existing work differs in how that forward pass is obtained, either by updating parameters on trajectory data (\S\ref{sec:learning-based}) or by shaping the input context of a frozen model (\S\ref{sec:prompt-based}). In the code-as-world-model paradigm (\S\ref{sec:programmatic}), the LLM is no longer the world model itself but its author: it emits executable code (PDDL, Python, HTML, etc.), and the world model is the code together with its executor. Figure~\ref{fig:building} traces the construction pipeline for each paradigm.

\subsection{Learning-Based Construction}
\label{sec:learning-based}

\subsubsection{Supervised Fine-Tuning on Trajectory Data}
\label{sec:sft-wm}

The most direct approach to building a text world model is to fine-tune an LLM on $(s_t, a_t, s_{t+1})$ tuples collected from environment interactions. The key design choices are \emph{what} to predict, \emph{where} the training data comes from and \emph{how much} data is needed.

\paragraph{Prediction targets: full states vs deltas}
The first design choice is what the world model should output given a state--action pair. Existing approaches fall into two categories: predicting the complete next state, or predicting only the change induced by an action.

\emph{Full-state prediction} ($\worldmodel: \mathcal{S} \times \mathcal{A} \to \mathcal{S}$) generates the entire next observation $s_{t+1}$. \citet{xie2024preconditioneffect} decomposes this into precondition prediction (what must hold for an action to apply) and effect prediction (how the state changes), operating over short natural-language world-state descriptions in commonsense action sequences. An empirical study by \citet{li2025wordtoworld} across five text environments shows that full-state prediction achieves single-step accuracy of $\sim$99\% on structured environments such as ALFWorld and SciWorld. In code environments, the Code World Model~\citep{faircwm2025} likewise adopts full-state prediction, generating complete execution outputs (stdout, return values, and termination status) given a program and its inputs. Full-state prediction is well-suited to environments with compact observation spaces, as well as settings that require direct, continuous interaction such as text games and code execution, where each step demands a complete, self-contained observation.

\input{figures/build-pipeline-v2}

\emph{Delta prediction} ($\worldmodel: \mathcal{S} \times \mathcal{A} \to \Delta(\mathcal{S})$) targets only the change induced by an action, reducing the output space and concentrating supervision on causally relevant information. This formulation is especially motivated by web environments, where observations (accessibility trees, HTML pages) span thousands of tokens yet a typical action modifies only a handful of DOM elements. Transition-focused observation abstraction~\citep{chae2025wma} extracts natural-language state-difference descriptions from raw HTML observations, then trains the world model to predict these compact deltas rather than full next-state pages. \citet{gu2025webdreamer} independently corroborate this design: ablations comparing simulation output formats find that natural-language state-change descriptions are competitive with raw HTML and accessibility-tree representations for scoring candidate actions at short horizons. The authors nonetheless caution against any strict-superiority claim, reporting that this format degrades fastest as planning horizons extend. DynaWeb~\citep{ding2026dynaweb} adopts a delta-centric target: the world model is trained to predict the natural-language state-change description $\Delta$, and the next accessibility tree is reconstructed at inference by applying the predicted delta, which focuses supervision on what changed while still supporting on-policy rollouts.

In summary, the choice of prediction target is tied to the observation space: compact, structured environments and short-output code settings~\citep{faircwm2025} favor full-state prediction for its simplicity and multi-turn rollout support, while large, redundant observation spaces (web pages, GUI screens) favor delta prediction for efficiency in test-time candidate scoring. Hybrid designs bridge both by generating deltas as a reasoning scaffold before producing complete observations.

\paragraph{Trajectory data collection}
Given a prediction target, the next question is where the training tuples come from. Existing sources span a spectrum of decreasing reality and decreasing cost. The most direct strategy deploys agents (expert models, the base model, or random explorers) in the real target environment and records the resulting $(s_t, a_t, s_{t+1})$ tuples~\citep{chae2025wma,ding2026dynaweb,wu2025rlvrworld}; Word2World~\citep{li2025wordtoworld} similarly relies on GPT-4o rollouts across five text environments, collecting 40k--70k trajectories each. SPA~\citep{chen2025spa} stays in the real environment but uses self-play rollouts, replacing hallucinated states with ground-truth observations to keep the distribution anchored. Scaling the real-environment route further, WebWorld~\citep{xiao2025webworld} harvests 1.06M real web trajectories through an automated three-level pipeline. Beyond this point, several methods sever the dependence on the target environment entirely. Simia~\citep{li2025simia} has an LLM simultaneously play user, agent, and environment from a handful of seed trajectories, synthesizing over 90k fully artificial transitions without any real environment access; \citet{naous2025flipping} repurposes 384k existing human conversations via dialogue role-flipping. CWM~\citep{faircwm2025} occupies a different niche, executing Python programs to record function-level traces, which is cheap to scale within the executable code domain but does not generalize beyond it.

Across this spectrum, reality and cost trade off monotonically: real rollouts give in-distribution evidence at high access cost, self-play and large-scale harvesting amortize that cost through automation, and fully synthetic or repurposed data drives cost to near zero at the price of being potentially less faithful to the target dynamics.

\paragraph{Data scale: from thousands to trillions}
Data requirements grow monotonically with environment openness, but the absolute scales span eight orders of magnitude. Word2World~\citep{li2025wordtoworld} observes clear scaling laws with distinct saturation behavior: closed, structured environments (ALFWorld, SciWorld) saturate at $\sim$20k trajectories, whereas open-ended environments (WebShop) continue to improve at 70k and tool-use environments (StableToolBench) remain unsaturated even at 160k. WebWorld~\citep{xiao2025webworld} scales the real-trajectory route to 1.06M web rollouts, training a 32B model that first learns transition dynamics on the full corpus and then activates reasoning with only 0.09\% chain-of-thought data. Pushing scale further still, the Code World Model (CWM;~\citealp{faircwm2025}) performs mid-training on 5T tokens enriched with executable code traces (interpreter outputs and Docker interaction logs), yielding a 32B model that learns to simulate program execution, predict outputs, and judge termination through continued pre-training on large-scale code execution traces, without requiring a live execution environment at inference time. This last result demonstrates that world-modeling competence can arise from continued pre-training at sufficient scale, mirroring the well-documented pattern in LLM research where new capabilities emerge as data and compute cross critical thresholds~\citep{wei2022emergentabilitieslargelanguage}.

\begin{takeaway}
The three design axes (what to predict, where the data comes from, how much is needed) are not independent. Delta prediction pays off only where the observation space is large and redundant, which is also where collection in the real environment is most expensive, hence its frequent pairing with web-trajectory pipelines. Fully synthetic or repurposed data is cheapest but inherits the LLM's existing coverage of the target dynamics, so it is most defensible when the prediction target is a short, local delta rather than a full state. And the data threshold itself shifts with environment openness: closed simulators saturate at $10^4$, open-ended web tasks demand $10^6$, and code execution requires continued pre-training at the trillion-token scale. Across all three axes, SFT remains bounded by the cost of obtaining ground-truth next states and by compounding error over long rollouts, motivating the RL-based alternatives discussed below.
\end{takeaway}

\subsubsection{Reinforcement Learning-Based Training}
\label{sec:rl-wm}

Whereas SFT optimizes token-level likelihood, RL-based training optimizes task-relevant rewards. The central design choice is therefore what property of a prediction the reward should measure: unlike agent RL, where task success provides a natural signal, world-model RL targets prediction quality, a notion without a single canonical definition. We organize existing rewards along a single axis, namely how far the supervision signal sits from the predicted token sequence and how close it sits to the prediction's downstream consequences. We focus here on rewards that target the world model itself, distinguishing them from two related uses of RL discussed in \S\ref{sec:training-time}: applying RL to a unified policy--world-model without a separate world-model loss (e.g., Dyna-Mind;~\citealp{yu2025dynamind}), and using RL solely for the policy objective (e.g., SPA;~\citealp{chen2025spa}), which are optimized with agent-centric task rewards rather than world-model-specific signals.

\paragraph{Surface fidelity}
The most direct reward compares the predicted next state to the actual state with a string-level metric. RLVR-World~\citep{wu2025rlvrworld} post-trains autoregressive world models with such metrics (exact match for text-game states, token-level F1 for web states) and reports consistent gains over an SFT-initialized baseline. Because the rewards are deterministically computable, they avoid reward hacking by construction, but surface matching cannot tell a semantically equivalent paraphrase from a factually wrong prediction.

\paragraph{Semantic equivalence}
A second family of rewards lifts the comparison from the literal string to its meaning, differing in who acts as the semantic judge. RWML~\citep{yu2025rwml} uses a frozen text encoder, defining a binary cosine-similarity reward in the encoder's embedding space, and pairs it with curriculum subsampling to position itself as a domain-agnostic mid-training stage that needs neither expert demonstrations nor task-success signals. CUWM~\citep{guan2026computerusingworldmodel} replaces the text encoder with an LLM-as-a-judge that scores each predicted transition along weighted UI structural aspects, with higher weights on decision-critical components, and drives GRPO optimization on GUI dynamics.
While these rewards introduce semantic flexibility, they still treat the predicted state as the object of interest, and a state that is semantically close may still lead to a different downstream decision.

\paragraph{Behavioral consistency}
A further shift moves the reward off state similarity entirely and onto whether the prediction would lead the agent to the same decision. BehR~\citep{huang2026stateconsistencybehaviorconsistency} identifies a metric inversion pathology: a predicted state can be both textually and semantically close to the ground truth yet drop the single decision-critical token (e.g., the target product in WebShop), while a textually divergent state that preserves that token still induces the correct action. BehR therefore evaluates a frozen reference policy on the logged next action under both the predicted and the true next state, and rewards the world model by the negative absolute log-likelihood gap. Optimized with GRPO, this signal preserves single-step exact match while substantially improving task-level consistency, indicating that aligning the reward with downstream behavior recovers what state-similarity rewards fail to capture.

\paragraph{Latent consistency}
The three preceding rewards each lean on external supervision: a recorded ground-truth state, a pretrained encoder or judge, or a reference agent. SWIRL~\citep{qiu2026selfimprovingworldmodellinglatent} removes this dependency by treating actions as latent variables and decomposing world modeling into a forward model and an inverse-dynamics model that alternate as policy and reward under GRPO, jointly maximizing a variational lower bound on next-state likelihood. Because actions appear only as latents, training consumes state-only sequences and scales to unlabeled corpora; the trade-off is that the learned latent actions carry no guarantee of aligning with the operational vocabulary of the downstream agent.

\begin{takeaway}
Across these four designs the reward target moves progressively further from the predicted token sequence and closer to its downstream consequences, but world-model RL remains noticeably less mature than its SFT counterpart. A key limitation is that surface, semantic, and latent rewards all focus on matching recorded observations, while a world model is used by an agent, so what really matters is whether predicted states lead to the same downstream decisions, as early work like BehR has highlighted.
\end{takeaway}

\subsection{Prompt-Based Construction}
\label{sec:prompt-based}

A complementary route stays within the LLM-as-world-model paradigm but forgoes parameter updates, leveraging the world knowledge already encoded in pretrained LLMs. We organize existing approaches by the source of the world knowledge they consume: the LLM's own prior alone (\S\ref{sec:incontext-wm}), a static external corpus (\S\ref{sec:retrieval-wm}), or experience accumulated through interaction (\S\ref{sec:self-evolving-wm}).

\subsubsection{In-Context World Modeling}
\label{sec:incontext-wm}

The most direct way to obtain a text world model without training is to treat a frozen LLM's forward pass as the transition function: conditioned on the current state $s_t$ and a candidate action $a_t$, the model is prompted to emit the next state $\hat{s}_{t+1}$. For web agents, WebDreamer~\citep{gu2025webdreamer} prompts GPT-4o to imagine a natural-language description of the resulting page and uses the imagined state to score short-horizon rollouts. For household tasks, LLM-MCTS~\citep{zhao2023llmmcts} elicits object-location beliefs from GPT-3.5 through repeated sampling and queries the same frozen model separately as a heuristic policy.

However, the reliability of such a setup is bounded by the LLM's intrinsic knowledge of the target domain: when the model lacks coverage of the relevant dynamics, prompting alone cannot fill the gap, and predictions degrade as the planning horizon grows. Empirically, LLMs exceed 75\% on next-state identification but rarely exceed 65\% on full-procedure planning alignment~\citep{mei2025rwom}, indicating that errors compound rapidly over multiple steps. The common limitation of this route is its inability to incorporate new knowledge into the world model, leaving the model's ceiling fixed by what the base LLM already knows.

\subsubsection{Retrieval-Augmented World Knowledge}
\label{sec:retrieval-wm}

Retrieval-augmented approaches mitigate the compounding-error problem by grounding the world model's predictions in external knowledge sources, so that each step is conditioned on relevant evidence rather than on the model's prior alone.

The first direction grounds the world model in procedural traces, retrieving tutorials or expert step sequences as conditioning context. R-WoM~\citep{mei2025rwom} retrieves and reranks tutorial passages and then runs a single-pass long-CoT rollout that imagines the full $k$-step trajectory in one reasoning call, instead of issuing $k\times m$ separate LLM calls for $m$ candidates; the grounded trajectory remains useful at horizons of up to three steps on OSWorld, beyond which performance degrades. TRAD~\citep{zhou2024trad} operates at the level of individual expert steps: it retrospectively annotates expert trajectories with LLM-generated ``thought'' abstractions and uses these thoughts as retrieval queries, with an alignment module that handles temporal mismatch when the retrieved step is reused.

The second direction compresses past experience into structured knowledge that is retrieved on demand. WorldMind~\citep{ren2025worldmind} maintains a natural-language knowledge base built from a Predict-Act-Verify loop, storing failure-induced feasibility constraints as causal rules (\eg, ``must hold a knife before slicing'') and distilling successful trajectories into procedural heuristics. Related systems vary the storage format: tuple-indexed experience~\citep{chari2025mindstores}, external affordances, or graph- and prototype-structured memories~\citep{chhikara2023knowledgetext, yang2025aec, yoo2025wormi}.

The choice between the two directions is a fidelity--generalization trade-off: retrieving full procedural traces gives concrete, in-distribution evidence at the cost of brittleness when the current task diverges, while retrieving distilled rules or prototypes generalizes more broadly but requires the distillation step to faithfully capture the relevant dynamics.
And both directions assume that useful prior experience already exists, which fails in cold-start environments and motivates the self-evolving designs.

\subsubsection{Self-Evolving Prompt World Models}
\label{sec:self-evolving-wm}

A third training-free approach lets the world model accumulate its knowledge through interaction. Instead of relying solely on a fixed corpus or a frozen prior, the agent records what happens during exploration or task execution, distills it into reusable knowledge, and feeds that knowledge back into subsequent prompts.

\citet{chen2025testtimeadapt} run a short pre-deployment exploration episode in a target environment using a small set of LLM-generated personas, distill the resulting state-transition triples into natural-language causal rules, and inject the rule set as a fixed in-context world model for all subsequent tasks. The cost is paid once per environment and the rule set then remains static at deployment, recovering much of the benefit of fine-tuning without any parameter updates. Steve-Evolving~\citep{xie2025steveevolving} instead keeps accumulating online during task execution: each Minecraft trial is paired with a fine-grained diagnosis, and the system writes back both positive macro skills (preconditions, action flow, effects) and negative guardrail rules that block known failure modes. The distilled knowledge is injected into the LLM planner's context on later tasks, and ablations show that the bottleneck is exposing this knowledge to the planner rather than merely storing it. Evo-Memory~\citep{wei2025evomemory} takes this idea further by treating memory maintenance as part of the agent's action space: the ReAct loop is extended with a Refine action that lets the agent reorganize past experiences, prune noisy entries, and promote reusable ones, so the world-knowledge store itself is updated by a learnable policy rather than a fixed distillation rule.

\subsection{Programmatic Construction: Code as World Model}
\label{sec:programmatic}

An increasingly prominent paradigm uses LLMs to generate executable code that serves as the world model. Unlike the LLM-as-world-model paradigm, where transitions are predicted by a forward pass, the code-as-world-model paradigm executes transitions deterministically, enabling formal verification, exact reproducibility, and a clean separation between the LLM's role (writing the code) and the world model's role (running it). The two questions that organize this subsection are what code is generated to act as the world model (\S\ref{sec:generated-envs}), and how such generation is scaled to large environment collections (\S\ref{sec:env-scaling}).

\subsubsection{What Code Is Generated}
\label{sec:generated-envs}

The form of the generated code varies by target domain, ranging from concrete renderers, to symbolic state machines, to full simulator code. As a concrete renderer, Code2World~\citep{zheng2025code2world} reframes GUI world modeling as renderable HTML generation, training a VLM to emit HTML that, when rendered, yields the next-state screenshot. As symbolic state machines, AutoWebWorld~\citep{wu2025autowebworld} models web environments as finite-state machines with deterministic, programmatically checkable transitions, while CLI-Gym~\citep{lin2025cligym} inverts a healthy CLI Docker image by issuing destructive commands to construct environment-modifying tasks. At the simulator level, Code World Models~\citep{lehrach2025codewm} compile natural-language game rules into a Python OpenSpiel implementation that supports MCTS planning, and TheoryCoder~\citep{ahmed2025theorycoder} extends this to bilevel planning where PDDL operators provide high-level structure and LLM-synthesized Python functions implement low-level transitions.

Across these works, what is being generated ranges from concrete renderers (HTML, screenshots) to symbolic state machines (FSMs, PDDL operators) to full simulator code, but the common move is to push the world model out of the LLM's forward pass and into an artifact that can be inspected, replayed, and verified independently.

\subsubsection{How to Scale Environment Synthesis}
\label{sec:env-scaling}

Once code-generated environments are cheap to produce, environment count itself becomes a scaling dimension for agent capability, alongside model size and trajectory volume. The open question is then less about writing more environments and more about keeping them useful: each must be functionally correct, and the collection must be diverse enough to prevent policy collapse.

\paragraph{Quality assurance and scaling evidence}
Pipelines combining automated verification with large-scale synthesis differ chiefly in the verification mechanism they rely on. EnvScaler~\citep{song2025envscaler} uses dual-agent verification (a testing agent probes edge cases; a checking agent inspects state changes) to produce 191 verified sandboxes with monotonic gains as environment count grows. ScaleEnv~\citep{tu2025scaleenv} validates synthesized tools via procedural unit tests and tool dependency graphs for compositional coverage. AWM~\citep{wang2025awm} backs each environment with a SQLite database so every tool call maps to a SQL query with verifiable pre/post-conditions, scaling to 1{,}000+ environments and improving GRPO training on BFCLv3 and OOD benchmarks. AgentScaler~\citep{fang2025agentscaler} clusters APIs into thousand-domain semantic tool graphs via Louvain detection, and daVinci-Env~\citep{fu2025davincienv} synthesizes 45K+ Docker environments from 10K+ repositories with difficulty-aware curation, yielding log-linear scaling on SWE-bench Verified.

\paragraph{Adaptive difficulty and diversity}
Beyond raw count, environment diversity is a complementary axis that determines whether scaling pays off. RLVE~\citep{zeng2025rlve} uses a sliding-window curriculum auto-incrementing difficulty when accuracy exceeds 90\%, establishing an environment-count scaling law across 400 environments. AutoEnv~\citep{zhang2025autoenv} generates rule-heterogeneous environments via a three-layer abstraction and shows that the gap between the best fixed strategy and an environment-adaptive upper bound widens with diversity. The principle extends to multimodal settings: Game-RL~\citep{tong2025gamerl} finds that game diversity directly determines OOD generalization on visual reasoning benchmarks.

\begin{takeaway}
Count and diversity are complements rather than alternatives: raw count gives log-linear gains, but those gains saturate quickly without diversity. The shared limitation of this line of work is that every verification mechanism is domain-specific (SQL pre/post-conditions, Docker exit codes, procedural unit tests), so there is no transferable standard for what counts as a ``correct'' synthesized environment, which keeps each pipeline siloed in its own domain.
\end{takeaway}

\subsection{Cross-Paradigm Comparison}
\label{sec:code-vs-neural}

\input{figures/foundations-spectrum}

\paragraph{When to use which paradigm}
The first question is whether the target dynamics admit a closed-form description. If transitions are fixed and expressible in code (game rules, tool APIs, GUI state machines), code-as-WM (\S\ref{sec:programmatic}) is the natural fit and yields deterministic, replayable, and verifiable transitions for free. When the dynamics are open-ended or hinge on broad world knowledge (everyday commonsense, long-tail web behavior), they cannot be hand-coded, and LLM-as-WM becomes necessary; within it, learning-based construction (\S\ref{sec:learning-based}) is preferred when sufficient trajectory data can be collected, while prompt-based construction (\S\ref{sec:prompt-based}) is more appropriate under limited data, leveraging the LLM's prior to fill the gap.

\paragraph{Strengths and limitations}
Figure~\ref{fig:foundations-spectrum} places the three constructions along an implicit-to-explicit spectrum of how world knowledge is represented. Learning-based methods achieve high fidelity and compression but sacrifice verifiability and long-horizon consistency. Prompt-based methods offer the lowest barrier to entry and rapid adaptation but suffer from hallucination and poor calibration. Code-as-WM methods provide deterministic, verifiable, and reusable transitions, at the cost of per-environment construction and a domain that admits a code-level specification.

\paragraph{Current trends}
Three trends are visible across the section. First, supervision is moving away from token-level fidelity: from SFT toward reward-based training within learning-based methods (\S\ref{sec:rl-wm}), and from a frozen prior toward retrieval and self-evolving knowledge stores within prompt-based methods. Second, scaling has shifted from collecting more trajectories to synthesizing more environments, with environment count and diversity emerging as first-class scaling axes (\S\ref{sec:env-scaling}). Third, the two top-level paradigms are not mutually exclusive in principle, and combining a code-grounded layer with an LLM-driven layer remains an underexplored direction.

%% file: figures/build-pipeline-v2.tex
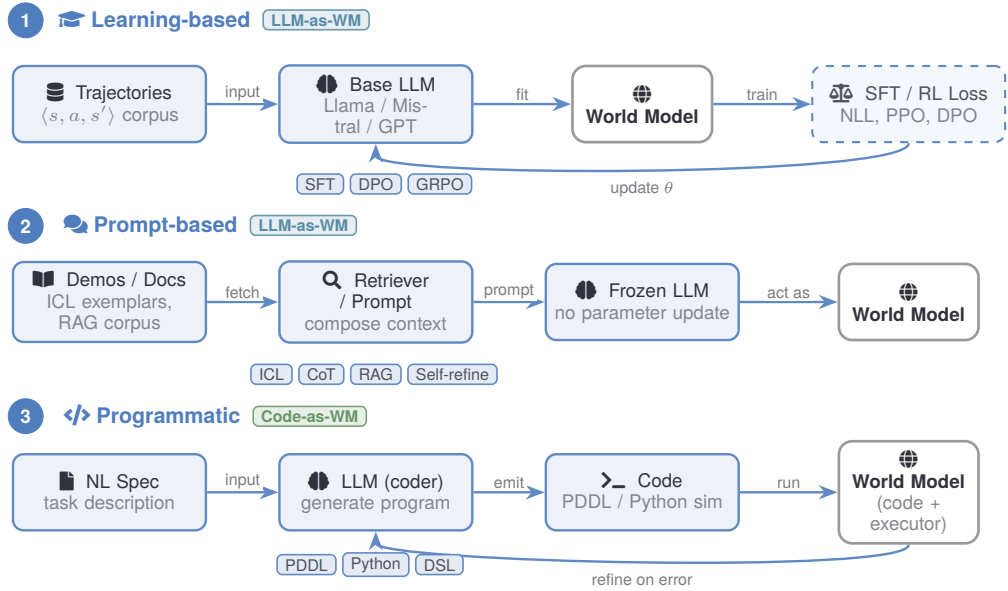
\begin{figure}
    \centering
    \resizebox{0.8\textwidth}{!}{%
    \begin{tikzpicture}[
        >=Stealth, font=\sffamily,
        numcircle/.style={circle, draw=morBuild, fill=morBuild, text=white,
            font=\sffamily\bfseries\fontsize{7}{8}\selectfont, inner sep=0pt, minimum size=13pt},
        rowtitle/.style={font=\sffamily\footnotesize\bfseries, text=morBuild, anchor=west},
        block/.style={rectangle, rounded corners=4pt,
            minimum width=2.4cm, minimum height=1.0cm, text width=2.3cm, align=center,
            font=\sffamily\scriptsize, text=morInk,
            draw=morBuild!90, fill=morBuild!10, line width=0.9pt},
        wmblock/.style={rectangle, rounded corners=5pt,
            minimum width=1.7cm, minimum height=1.0cm, text width=1.6cm, align=center,
            font=\sffamily\scriptsize\bfseries, text=morInk,
            draw=morInk!50, fill=white, line width=1pt},
        outblock/.style={rectangle, rounded corners=4pt,
            minimum width=2.4cm, minimum height=1.0cm, text width=2.3cm, align=center,
            font=\sffamily\scriptsize, text=morInk!85,
            draw=morBuild!90, fill=morBuild!6, line width=0.8pt, dashed},
        chip/.style={rectangle, rounded corners=2pt, draw=morBuild!90, fill=morBuild!15,
            font=\sffamily\fontsize{6}{7}\selectfont, text=morInk!85,
            inner xsep=3pt, inner ysep=1.5pt, line width=0.6pt},
        tag/.style={rectangle, rounded corners=2pt, draw=#1!90, fill=#1!15,
            font=\sffamily\fontsize{6}{7}\selectfont\bfseries, text=#1,
            inner xsep=3pt, inner ysep=1.5pt, line width=0.6pt},
        arr/.style={->, line width=0.9pt, color=morBuild!90},
        lbl/.style={font=\sffamily\fontsize{6}{7}\selectfont, text=morInk!65, fill=white, inner sep=1pt},
    ]
    \node[numcircle] (n1) at (-1.1, 0) {1};
    \node[rowtitle, right=2pt of n1] {\faGraduationCap~Learning-based~~\tikz[baseline=-0.6ex]\node[tag=morAgent] {LLM-as-WM};};
    \node[block]   (l1) at (0, -1.1) {\faDatabase~~Trajectories\\\textcolor{morInk!60}{$\langle s,a,s'\rangle$ corpus}};
    \node[block]   (l2) at (3.5, -1.1) {\faBrain~~Base LLM\\\textcolor{morInk!60}{Llama / Mistral / GPT}};
    \node[wmblock] (l3) at (7.0, -1.1) {\faGlobe\\World Model};
    \node[outblock](l4) at (10.5, -1.1) {\faBalanceScale~~SFT / RL Loss\\\textcolor{morInk!60}{NLL, PPO, DPO}};
    \draw[arr] (l1) -- node[lbl, above] {input} (l2);
    \draw[arr] (l2) -- node[lbl, above] {fit} (l3);
    \draw[arr] (l3) -- node[lbl, above] {train} (l4);
    \draw[arr] (l4.south) .. controls +(0,-0.45) and +(0,-0.45) .. node[lbl, below, yshift=-3pt] {update $\theta$} (l2.south);
    \node[chip] (l-c4) at (3.5, -2.15) {DPO};
    \node[chip, left=2pt of l-c4] (l-c1) {SFT};
    \node[chip, right=2pt of l-c4] (l-c5) {GRPO};

    \node[numcircle] (n2) at (-1.1, -2.7) {2};
    \node[rowtitle, right=4pt of n2] {\faComments~Prompt-based~~\tikz[baseline=-0.6ex]\node[tag=morAgent] {LLM-as-WM};};
    \node[block]   (p1) at (0, -3.7) {\faBookOpen~~Demos / Docs\\\textcolor{morInk!60}{ICL exemplars,\\RAG corpus}};
    \node[block]   (p2) at (3.5, -3.7) {\faSearch~~Retriever / Prompt\\\textcolor{morInk!60}{compose context}};
    \node[block]   (p3) at (7.0, -3.7) {\faBrain~~Frozen LLM\\\textcolor{morInk!60}{no parameter update}};
    \node[wmblock] (p4) at (10.5, -3.7) {\faGlobe\\World Model};
    \draw[arr] (p1) -- node[lbl, above] {fetch} (p2);
    \draw[arr] (p2) -- node[lbl, above] {prompt} (p3);
    \draw[arr] (p3) -- node[lbl, above] {act as} (p4);
    \node[chip] (p-c3) at (3.5, -4.65) {RAG};
    \node[chip, left=2pt of p-c3] (p-c2) {CoT};
    \node[chip, left=2pt of p-c2] (p-c1) {ICL};
    \node[chip, right=2pt of p-c3] (p-c4) {Self-refine};

    \node[numcircle] (n3) at (-1.1, -5.2) {3};
    \node[rowtitle, right=4pt of n3] {\faCode~Programmatic~~\tikz[baseline=-0.6ex]\node[tag=morEnv] {Code-as-WM};};
    \node[block]   (g1) at (0, -6.2) {\faFile~~NL Spec\\\textcolor{morInk!60}{task description}};
    \node[block]   (g2) at (3.5, -6.2) {\faBrain~~LLM (coder)\\\textcolor{morInk!60}{generate program}};
    \node[block]   (g3) at (7.0, -6.2) {\faTerminal~~Code\\\textcolor{morInk!60}{PDDL / Python sim}};
    \node[wmblock] (g4) at (10.5, -6.2) {\faGlobe\\World Model\\\textmd{\textcolor{morInk!60}{(code + executor)}}};
    \draw[arr] (g1) -- node[lbl, above] {input} (g2);
    \draw[arr] (g2) -- node[lbl, above] {emit} (g3);
    \draw[arr] (g3) -- node[lbl, above] {run} (g4);
    \draw[arr] (g4.south) .. controls +(0,-0.4) and +(0,-0.4) .. node[lbl, below, yshift=-3pt] {refine on error} (g2.south);
    \node[chip] (g-c2) at (3.5, -7.15) {Python};
    \node[chip, left=2pt of g-c2] (g-c1) {PDDL};
    \node[chip, right=2pt of g-c2] (g-c3) {DSL};

    \end{tikzpicture}}
    \caption{Construction pipelines for the three building paradigms (\S\ref{sec:building}). Each row traces the data flow from input to a usable world model: \textbf{(1) Learning-based} feeds $\langle s,a,s'\rangle$ trajectories into a base LLM and updates parameters via SFT, DPO, or GRPO losses. \textbf{(2) Prompt-based} composes demonstrations or retrieved documents into a context that turns a frozen LLM into a world model via ICL, CoT, RAG, or self-refine. \textbf{(3) Programmatic} prompts a coder LLM to emit PDDL, Python, or DSL programs that an executor runs as the world model, with execution errors fed back for refinement.}
    \label{fig:building}
\end{figure}

%% file: figures/foundations-spectrum.tex
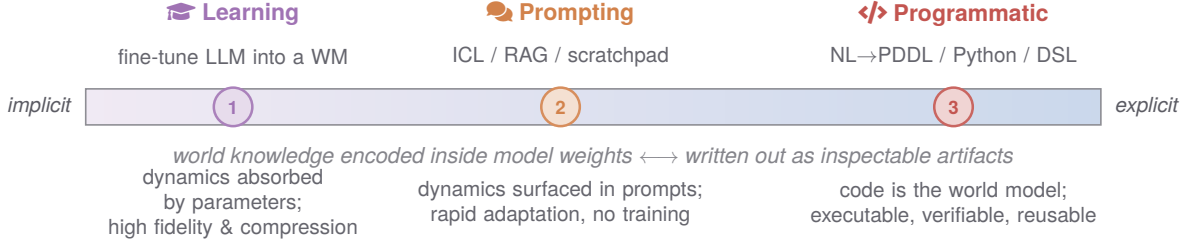
\begin{figure}
    \centering
    \resizebox{0.95\textwidth}{!}{%
    \begin{tikzpicture}[
        >=Stealth, font=\sffamily,
        bar/.style={rectangle, draw=morInk!55, line width=0.9pt, minimum height=0.55cm,
            minimum width=15.5cm, anchor=west},
        marker/.style={circle, draw=#1!90, fill=#1!25, line width=1.1pt,
            minimum size=0.55cm, font=\sffamily\bfseries\footnotesize, text=#1},
        cap/.style={font=\sffamily\normalsize\bfseries, text=#1, align=center},
        sub/.style={font=\sffamily\fontsize{9}{10.8}\selectfont, text=morInk!75, align=center, text width=4.8cm},
        axislbl/.style={font=\sffamily\fontsize{9}{10.5}\selectfont\itshape, text=morInk!80},
        arr/.style={->, line width=1.0pt, color=morInk!55},
    ]
    \shade[left color=morFound!15, right color=morBuild!30]
        (-7.75, 0) rectangle (7.75, 0.55);
    \draw[morInk!55, line width=0.7pt] (-7.75, 0) rectangle (7.75, 0.55);

    \node[axislbl, anchor=east] at (-7.85, 0.275) {implicit};
    \node[axislbl, anchor=west] at ( 7.85, 0.275) {explicit};
    \node[font=\sffamily\fontsize{9.5}{11}\selectfont, text=morInk!60] at (0, -0.5)
        {\textit{world knowledge encoded inside model weights $\longleftrightarrow$ written out as inspectable artifacts}};

    \node[marker=morFound] (m1) at (-5.5, 0.275) {1};
    \node[marker=morTrain] (m2) at (-0.5, 0.275) {2};
    \node[marker=morEval]  (m3) at ( 5.5, 0.275) {3};

    \node[cap=morFound] at (-5.5, 1.7) {\faGraduationCap~Learning};
    \node[sub] at (-5.5, 1.05) {fine-tune LLM into a WM};

    \node[cap=morTrain] at (-0.5, 1.7) {\faComments~Prompting};
    \node[sub] at (-0.5, 1.05) {ICL / RAG / scratchpad};

    \node[cap=morEval] at ( 5.5, 1.7) {\faCode~Programmatic};
    \node[sub] at ( 5.5, 1.05) {NL$\rightarrow$PDDL / Python / DSL};

    \node[sub] at (-5.5, -1.2)
        {dynamics absorbed by parameters;\\high fidelity \& compression};
    \node[sub] at (-0.5, -1.2)
        {dynamics surfaced in prompts;\\rapid adaptation, no training};
    \node[sub] at ( 5.5, -1.2)
        {code is the world model;\\executable, verifiable, reusable};

    \end{tikzpicture}}
    \caption{The three construction paradigms of \S\ref{sec:building} along an implicit$\leftrightarrow$explicit spectrum of how world knowledge is represented: \textbf{learning} absorbs dynamics into model parameters (high fidelity, low verifiability); \textbf{prompting} surfaces dynamics via in-context exemplars or retrieval (rapid adaptation, hallucination-prone); \textbf{programmatic} synthesis emits executable artifacts (verifiable and reusable, but per-environment).}
    \label{fig:foundations-spectrum}
\end{figure}

%% file: sections/04-training-time.tex
\section{Training-Time World Models}
\label{sec:training-time}

Having surveyed how text world models are constructed (\S\ref{sec:building}), we now turn to how they are used to improve agents before deployment. Whereas the previous section took the world model itself as the object of study, this section shifts perspective to the agent as the primary beneficiary, and asks what role the world model plays inside the training loop. Three roles emerge (Figure~\ref{fig:training-loops}): the world model can be folded into the agent's own parameters so that anticipation travels with the policy (\S\ref{subsec:internalized-wm}); it can act as an external substitute for the system environment, providing observations and rewards instead of a real testbed (\S\ref{subsec:llm-env-substrate}); or it can simulate a human user, supplying interactive partners whose dynamics differ qualitatively from those of system environments (\S\ref{subsec:user-sim-training}).

\input{figures/training-loops-v3}

\subsection{Internalizing World Models into Agent Parameters}
\label{subsec:internalized-wm}

Action selection is inherently anticipatory: choosing well requires some expectation of what each candidate action will bring about. For LLM agents deployed in unfamiliar environments, pretraining alone does not supply this expectation; acquiring it through trial-and-error is expensive at training time, and external simulators are often unavailable at deployment. One response is to fold environment dynamics directly into the agent's parameters, so that anticipation travels with the policy rather than living in a separate module. This brings three benefits, namely no additional inference call, co-adaptation of world-modeling and decision-making representations, and transfer to settings where no simulator exists at test time. Existing work then divides on a further question of when the internalized world model exerts its influence: one line treats world modeling purely as a warmup signal whose effect persists implicitly in the weights, while the other surfaces state predictions in the reasoning trace and makes simulation an explicit step in action selection.

\subsubsection{World model as warm-start}
The first thread keeps the internalized world model implicit in the weights, motivated by the concern that retaining a world-modeling loss alongside the policy objective causes the two objectives to interfere; the shared move is to decouple world-model training from policy optimization in time. SPA~\citep{chen2025spa} learns transition dynamics and state representations through self-play SFT, after which PPO-style optimization proceeds without any auxiliary world-modeling loss. Early Experience~\citep{zhang2025earlyexperience} extends the same decoupling to imitation learning: two reward-free auxiliary objectives, implicit next-state prediction, and natural-language self-reflection on sub-optimal alternative actions, pre-train a checkpoint that is then fine-tuned on expert trajectories, and the same checkpoint also serves as a stronger warm-start for optional downstream RL. RWML~\citep{yu2025rwml} extends the warm-start view by training the world-modeling stage with GRPO under a binarized cosine-similarity reward in pretrained embedding space, sidestepping both the brittleness of token-level matching and the reward hacking invited by LLM-as-judge scoring; a second GRPO stage then optimizes for task success, and the RL-based world-modeling phase is reported to induce substantially less catastrophic forgetting than an SFT counterpart.

\subsubsection{World model in the reasoning trace}
The second thread instead surfaces the world model at decision time: since it lives in the agent's parameters, its predictions can also be exposed during inference, allowing the agent to consult them when choosing actions. Dyna-Think~\citep{yu2025dynathink} trains the model to predict next states for candidate actions inside the reasoning trace and to commit to the action whose simulated outcome best advances the task; among three world-modeling objectives compared (next-state prediction, state-change prediction, and teacher-generated critiques contrasting simulated with actual transitions), the critique variant performs best. Dyna-Mind~\citep{yu2025dynamind} carries this idea into online RL: a first stage distils real-environment search trees into reasoning chains exhibiting ``simulate $\to$ compare $\to$ decide'' patterns through SFT, while a second stage executes the agent's imagined plan in the real environment and feeds ground-truth next states back as text supervision, jointly optimizing simulation accuracy and task success. The authors report a strong positive correlation between simulation quality and downstream task success, indicating that the reasoning trace and the implicit world model improve in tandem.

The two threads localize the payoff of internalization at different points in the agent loop: keeping the world model as a warm-start confines its influence to training, yielding cleaner optimization and shorter inference at the cost of foregoing simulation at action time, while surfacing it in the reasoning trace couples simulation with action selection at the cost of additional optimization machinery and inference-time tokens.

\subsection{World Models as Training Environments}
\label{subsec:llm-env-substrate}

A further line of work drops the real testbed entirely and trains the agent on data fabricated by a world model. Whereas \S\ref{sec:building} catalogued how such environments are constructed, this section examines how they are consumed during agent training. We organize approaches by the coupling between the world model and the training loop, from weakest to strongest: a one-shot offline data source (\S\ref{subsubsec:offline-synth}), an online environment that responds on every rollout step (\S\ref{subsubsec:online-llm-env}), and a co-evolving partner that is updated alongside the policy (\S\ref{subsubsec:coevol-wm}).

\subsubsection{Offline Trajectory Synthesis}
\label{subsubsec:offline-synth}

At the weakest end of the coupling spectrum, offline synthesis treats the world model as a one-shot data source: trajectories are generated, filtered, and then handed to a downstream SFT or behavior-cloning stage, with no further interaction. WebSynthesis~\citep{gao2025websynthesis} pairs an LLM-based world model with MCTS to explore action trees in virtual web environments, distilling both successful paths and failure-recovery rollback trajectories; the resulting BC agent matches a baseline trained on a comparable amount of real-environment data on WebArena-Lite. Simia-SFT~\citep{li2025simia} adopts the same data-amplification view: a single LLM pass that fabricates the entire query--tool-call--response chain inflates a small seed corpus into an order-of-magnitude larger synthetic one, without any deployed environment. At a coarser granularity, AgentScaler~\citep{fang2025agentscaler} produces an offline trajectory corpus by random-walking on its tool dependency graphs, and shows that a mid-sized model trained on this corpus matches much larger baselines on function-calling benchmarks. Word2World~\citep{li2025wordtoworld} reports a similar finding at smaller scale: SFT on as few as 1K WM-generated trajectories matches the same volume of real-environment data.

\subsubsection{Online WM-as-Environment}
\label{subsubsec:online-llm-env}

When the world model remains in the loop during RL, it returns observations and rewards on every step of a rollout, so the agent's own exploration shapes the data it learns from. This on-policy mode is more expensive than offline synthesis but closes the distribution gap between training data and the agent's actual behavior. DreamGym~\citep{chen2025dreamgym} runs fully online RL against a lightweight experience model that predicts next states and rewards on the fly, paired with a curriculum task generator, achieving non-trivial gains on WebArena-Lite with zero real-environment access. Simia-RL~\citep{li2025simia} uses the same LLM simultaneously as environment simulator and reward calculator, enabling GRPO without deploying any real environment; counterintuitively, RL on simulated environments outperforms RL on real ones on OfficeBench, suggesting that LLM-simulated dynamics provide more consistent and explorable training signals than noisy real environments. DeepAgent~\citep{li2026deepagent} applies the same recipe to tool-calling agents, replacing real API calls with an LLM-based tool simulator during RL training, with consistent gains over the CodeAct baseline on ToolBench and WebShop. SPICE~\citep{liu2025spice} converts unlabeled web documents into a verifiable RL environment via self-play, with a single LLM alternating between a Challenger that sees a document and writes grounded questions and a Reasoner that must answer without it; information asymmetry prevents symmetry collapse and document grounding blocks hallucination amplification.

A recurring failure mode of the pure online setting is hallucination drift: as the agent explores states the simulator was never trained on, the simulator's responses diverge from any real environment. Existing work addresses this along two complementary directions. The first re-grounds the simulator in structured state, as in AWM~\citep{wang2025awm}, where each synthesized environment is backed by a SQLite database whose tool calls map to verifiable SQL queries. The second restricts the simulator's output space, as in CUWM~\citep{guan2026computerusingworldmodel} and Code2World~\citep{zheng2025code2world}, which constrain GUI dynamics to textual transitions plus diffusion rendering or to renderable HTML; both inherit construction details from \S\ref{sec:programmatic} and yield more stable rollouts when used as the agent's RL environment than open-ended LLM simulation.

\subsubsection{Co-Evolving the Agent and the World Model}
\label{subsubsec:coevol-wm}

At the strongest end of the coupling spectrum, the world model itself evolves during agent training, so the simulator improves alongside the policy it serves rather than remaining frozen at construction time. DynaWeb~\citep{ding2026dynaweb} instantiates a full on-policy MBRL framework for web agents: the world model and the policy are updated inside the same RL loop, with rollouts mixing imagined and a small fraction of real expert trajectories, showing that an LLM-based world model can stably participate in on-policy RL. WebEvolver~\citep{fang2025webevolver} realizes co-evolution in a looser, iterative form: instead of a single RL loop, it alternates rejection-sampling SFT rounds in which the policy and the world model are jointly fine-tuned on successful trajectories, and the updated world model then synthesizes new imagined webpage observations on which the next round of agent training is performed, breaking the self-improvement plateau of frozen-WM agents on Mind2Web-Live, WebVoyager, and GAIA-web.

Tightening the coupling between world model and training loop trades training cost for distribution alignment: offline synthesis is cheapest but exposes the agent to a fixed snapshot of the simulator's coverage, online interaction closes the distribution gap at the price of running the simulator inside every rollout, and co-evolution further closes the loop by letting the simulator track the policy, but at the cost of joint stability concerns and rare empirical evidence beyond web agents. The shared limitation across all three regimes is hallucination drift: free-form LLM simulators degrade as the agent explores out-of-distribution states, and current mitigations either fall back to structured state (databases, code) or borrow constrained output spaces from \S\ref{sec:programmatic}, so the question of how to keep an open-ended LLM simulator faithful under on-policy exploration remains open.

\subsection{User Simulation for Agent Training}
\label{subsec:user-sim-training}

The environments discussed in \S\ref{subsec:llm-env-substrate} simulate system dynamics: page transitions, API responses, and terminal outputs. Simulating a human user introduces additional difficulty: users are stochastic, preference-driven, and underspecified in their communication, requiring distinct modeling choices and evaluation criteria. We therefore treat user simulation as a separate training paradigm.

\subsubsection{RL with Simulated User Environments}
\label{subsubsec:user-rl}

One line of work uses LLM-simulated users as interactive RL environments. Existing methods can be ordered by the complexity of the simulated user, from cooperative task-oriented users to vague, emotional, and persona-driven ones, and reward design in turn grows from a single task-success signal to dedicated reward models that capture each new layer of user behavior.

UserRL~\citep{qian2025userrl} provides a systematic framework with standardized gym environments spanning intent clarification, persuasion, travel planning, and tool-calling, all driven by LLM-simulated users. It studies reward shaping along two orthogonal dimensions, turn-level and trajectory-level, with an SFT cold-start to prevent early collapse; the optimal combination enables an open-source mid-sized model to surpass proprietary baselines, confirmed by real human tests. Real users, however, are often vague rather than cooperative. \citet{sun2025proactive} address this by simulating ambiguous users through prompt vaguenization, which auto-rewrites precise specifications into underspecified prompts, and pair this with a preference-aware simulator; productivity, proactivity, and personalization are jointly optimized via GRPO, and the resulting agent learns to ask clarifying questions only when necessary.

Moving beyond task-oriented interactions, users also exhibit emotional responses that agents must handle appropriately. Echo-N1~\citep{zhang2025echon1} trains dedicated humanlike and empathy reward models, both using discrete 0/1 outputs with chain-of-thought reasoning to resist reward hacking, enabling a mid-sized model to substantially outperform much larger commercial character models on the EPM-Index. A further step toward realism is simulating users with persistent personas. HER~\citep{du2026her} introduces dual-layer thinking that separates implicit system planning, hidden from the user, from explicit in-character monologue, and pairs this with a principle-aligned generative reward model at near-human agreement.

\subsubsection{User-Model Fidelity and Personalization}
\label{subsubsec:personalization}

The RL paradigm above assumes the simulated user faithfully proxies real humans; when this assumption fails, the agent overfits to simulator artifacts. This raises three dependent questions: how faithful are current user models, how can agents adapt to individual users given a faithful model, and can agents eventually learn from real users directly?

\paragraph{User-model fidelity}
\citet{naous2025flipping} train a dedicated UserLM-8b on 384k real human--assistant dialogues and find that agent success drops sharply when it replaces a prompted GPT-4o user, exposing weaknesses masked by overly cooperative simulators. The systematic overestimation of agent competence by prompted assistant LLMs motivates dedicated user-model training, and bears directly on evaluation (\S\ref{subsec:wm-as-eval}). HumanLM~\citep{wu2026humanlmsimulatingusersstate} addresses the problem from the modeling side: instead of imitating surface conversational patterns, it models users' latent psychological states (beliefs, goals, emotion, communication style) and aligns via GRPO. A blind human study finds its responses closest to participants' own, suggesting that state-level modeling is a more effective inductive bias for user fidelity than response imitation.

\paragraph{Agent-side adaptation}
Given a faithful user model, the bottleneck shifts to the agent's ability to adapt to individual preferences. PAHF~\citep{liang2026pahf} proposes dual-channel feedback with explicit per-user memory: pre-action clarification to detect ambiguity and post-action correction to update stale beliefs. Ablations show that neither channel alone is sufficient; experiments across embodied manipulation and online shopping confirm that the combination adapts to preference drift with substantially lower cumulative error. For cold-start settings where no history exists, Pep~\citep{bose2026pep} decomposes personalization into offline structure learning over population preferences and online Bayesian query selection, indicating that inference structure, rather than model capacity, is the binding constraint in this regime.

\paragraph{From simulated to real users}
The methods above still operate within user world models. OpenClaw-RL~\citep{wang2026openclawrl} moves beyond simulation entirely, learning directly from live interactions via process reward model judgments and hindsight-guided on-policy distillation on a fully asynchronous architecture.

\begin{takeaway}
The progression from simulation-only training, through fidelity-aware user modeling, to direct online learning from real users mirrors the broader sim-to-real trajectory of this section and resurfaces in \S\ref{sec:open-problems}. The shared limitation across the simulation-based methods is that fidelity is benchmarked against either prompted LLM users or LLM-as-judge scores, both of which are themselves simulators of the quantity they claim to measure. Any reported improvement therefore risks reflecting alignment with the evaluator rather than with real users. Closing this loop requires either dedicated user models trained on human dialogues (\citealp{naous2025flipping}) or direct evaluation against participants, both of which remain rare.
\end{takeaway}

\subsection{Summary and Comparative Analysis}
\label{subsec:training-time-summary}

\begin{table}[t]
\centering
\small
\caption{Comparison of training-time world model approaches. \textbf{WM Form}: how the world model manifests; \textbf{Real Env.}: whether real environment access is needed during training.}
\label{tab:training-time-comparison}
\resizebox{0.8\columnwidth}{!}{%
\begin{tabular}{@{}llcc@{}}
\toprule
\textbf{Method} & \textbf{Paradigm} & \textbf{WM Form} & \textbf{Real Env.} \\
\midrule
SPA~\citep{chen2025spa} & Internalized & Internal & Yes \\
Early Exp.~\citep{zhang2025earlyexperience} & Internalized & Internal & Yes \\
RWML~\citep{yu2025rwml} & Internalized & Internal & Yes \\
Dyna-Think~\citep{yu2025dynathink} & Internalized & Internal & Yes \\
Dyna-Mind~\citep{yu2025dynamind} & Internalized & Internal & Yes \\
\midrule
WebSynthesis~\citep{gao2025websynthesis} & WM-Env (offline) & External LLM & No \\
Simia-SFT~\citep{li2025simia} & WM-Env (offline) & LLM-simulated & No \\
AgentScaler~\citep{fang2025agentscaler} & WM-Env (offline) & Code env & No \\
DreamGym~\citep{chen2025dreamgym} & WM-Env (online) & Experience LLM & No \\
Simia-RL~\citep{li2025simia} & WM-Env (online) & LLM-simulated & No \\
DeepAgent~\citep{li2026deepagent} & WM-Env (online) & LLM-simulated & No \\
SPICE~\citep{liu2025spice} & WM-Env (online) & Documents & No \\
DynaWeb~\citep{ding2026dynaweb} & WM-Env (co-evol) & External LLM & Partial \\
WebEvolver~\citep{fang2025webevolver} & WM-Env (co-evol) & External LLM & Partial \\
\midrule
UserRL~\citep{qian2025userrl} & User Sim & LLM-simulated & No \\
Proactive~\citep{sun2025proactive} & User Sim & LLM-simulated & No \\
Echo-N1~\citep{zhang2025echon1} & User Sim & LLM-simulated & No \\
HER~\citep{du2026her} & User Sim & LLM-simulated & No \\
HumanLM~\citep{wu2026humanlmsimulatingusersstate} & User Sim & Trained user model & No \\
PAHF~\citep{liang2026pahf} & User Sim & Prompting+memory & No \\
Pep~\citep{bose2026pep} & User Sim & Bayesian + offline & No \\
OpenClaw-RL~\citep{wang2026openclawrl} & User Sim & Live users & Yes \\
\bottomrule
\end{tabular}%
}
\vspace{-10pt}
\end{table}

\paragraph{Three roles in the training loop}
The three paradigms address distinct questions about the training loop. Internalization (\S\ref{subsec:internalized-wm}) addresses how an agent retains environment knowledge after training, so that anticipation does not require an external simulator at deployment. World-model-as-environment (\S\ref{subsec:llm-env-substrate}) addresses how training proceeds when real-environment access is limited or costly, substituting a simulator that ranges from one-shot offline data to a co-evolving partner. User simulation (\S\ref{subsec:user-sim-training}) addresses how human dynamics are incorporated into the training loop, where real users would otherwise be too stochastic or sparse for RL.

\paragraph{Cross-cutting failure modes}
Several failure modes recur across the section. The sim-to-real gap is the most visible: training competence does not transfer automatically to deployment, and the gap is widest for user simulation, where prompted assistant LLMs systematically overestimate agent ability~\citep{naous2025flipping, zhou2026mindsim2realgapuser}. Coverage drift arises as the policy moves into states the simulator was not built for, with simulated responses diverging from any real environment.

\paragraph{Trends}
World models are evolving from static, frozen assets into dynamic partners. Offline trajectories are being replaced by on-policy rollouts (\S\ref{subsubsec:online-llm-env}), and increasingly by joint co-evolution with the policy (\S\ref{subsubsec:coevol-wm}). In parallel, user modeling is moving away from prompt-based assistant LLMs toward dedicated user models trained on human dialogue data, a necessary step toward evaluation pipelines that remain independent of the training source.

%% file: figures/training-loops-v3.tex
\begin{figure}
    \centering
    \resizebox{0.99\textwidth}{!}{%
    \begin{tikzpicture}[
        >=Stealth, font=\sffamily,
        panel/.style={rectangle, rounded corners=5pt,
            draw=morTrain!90, fill=morTrain!8, line width=1pt,
            minimum width=6.8cm, minimum height=3.9cm},
        ptitle/.style={font=\sffamily\normalsize\bfseries, text=morTrain, anchor=south west},
        wm/.style={circle, draw=morInk!50, fill=white, line width=1pt,
            minimum size=0.95cm, font=\sffamily\fontsize{8}{9}\selectfont\bfseries, text=morInk, align=center},
        ag/.style={circle, draw=morAgent!90, fill=morAgent!10, line width=1pt,
            minimum size=0.95cm, font=\sffamily\fontsize{8}{9}\selectfont\bfseries, text=morInk, align=center},
        wmag/.style={circle, draw=morAgent!90, fill=morAgent!10, line width=1.2pt,
            minimum size=1.4cm, font=\sffamily\fontsize{9}{10}\selectfont\bfseries, text=morInk, align=center},
        data/.style={rectangle, rounded corners=4pt,
            draw=morTrain!90, fill=morTrain!10, line width=0.9pt,
            minimum width=1.8cm, minimum height=0.8cm,
            font=\sffamily\fontsize{10}{11}\selectfont, text=morInk, align=center},
        arr/.style={->, line width=1.0pt, color=morTrain!90},
        lbl/.style={font=\sffamily\fontsize{7.5}{8.5}\selectfont, text=morInk!70, inner sep=1.5pt},
    ]
    \node[panel] (P1) at (-7.0, 0) {};
    \node[ptitle] at ([xshift=4pt, yshift=2pt]P1.north west) {\faBrain~\textcircled{\scriptsize 1}~Internalized};
    \node[wmag] (P1c) at ($(P1.center)+(-0.6, 0.0)$) {\faGlobe\,\faRobot\\WM\,$=\pi$};
    \draw[arr] (P1c.north east) .. controls +(1.4,0.7) and +(1.4,-0.7) ..
        node[lbl, right, xshift=2pt] {shared $\theta$} (P1c.south east);

    \node[panel] (P2) at (0, 0) {};
    \node[ptitle] at ([xshift=4pt, yshift=2pt]P2.north west) {\faMoon~\textcircled{\scriptsize 2}~WM-as-Env};
    \node[wm]   (P2wm)   at ($(P2.center)+(-2.4, 0.45)$) {\faGlobe\\WM};
    \node[data] (P2data) at ($(P2.center)+( 0.0, 0.45)$) {\faCloud~rollouts};
    \node[ag]   (P2ag)   at ($(P2.center)+( 2.4, 0.45)$) {\faRobot\\agent};
    \draw[arr] (P2wm)   -- node[lbl, above] {imagine} (P2data);
    \draw[arr] (P2data) -- node[lbl, above] {train} (P2ag);
    \draw[arr] (P2ag.south) .. controls +(0,-1.1) and +(0,-1.1) ..
        node[lbl, below, yshift=-4pt] {action} (P2wm.south);

    \node[panel] (P3) at (7.0, 0) {};
    \node[ptitle] at ([xshift=4pt, yshift=2pt]P3.north west) {\faUserFriends~\textcircled{\scriptsize 3}~User Simulation};
    \node[wm]   (P3wm)  at ($(P3.center)+(-2.4, 0.45)$) {\faUser\\WM};
    \node[data] (P3rew) at ($(P3.center)+( 0.0, 0.45)$) {\faComment~dialog\\\faStar~reward};
    \node[ag]   (P3ag)  at ($(P3.center)+( 2.4, 0.45)$) {\faHeadset\\asst.};
    \draw[arr] (P3wm)  -- node[lbl, above] {utter.} (P3rew);
    \draw[arr] (P3rew) -- node[lbl, above] {resp.} (P3ag);
    \draw[arr] (P3ag.south) .. controls +(0,-1.1) and +(0,-1.1) ..
        node[lbl, below, yshift=-4pt] {update} (P3wm.south);
    \end{tikzpicture}}
    \caption{Three training-time paradigms that pair a world model with an agent: \textcircled{\scriptsize 1} internalising the world model into the agent's own parameters, \textcircled{\scriptsize 2} using a world model as a training environment (offline synthesis, online rollouts, or co-evolution), and \textcircled{\scriptsize 3} simulating users for multi-turn interaction.}
    \label{fig:training-loops}
\end{figure}
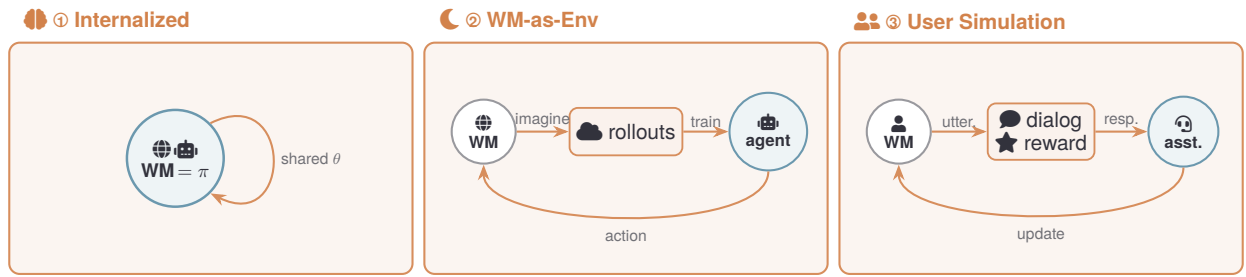

%% file: sections/05-inference-time.tex
\section{Inference-Time World Models}
\label{sec:inference-time}

While \S\ref{sec:training-time} examined how world models improve agents before deployment, this section addresses the complementary question of how text world models guide agent behavior at test time. The shared insight is that a world model lets the agent look ahead, simulating the consequences of candidate actions before committing to execution in the real environment. We organize inference-time uses by the role the world model plays: as a simulator that produces candidate futures to drive action selection (\S\ref{subsec:wm-as-simulator}), or as a verifier that screens or rewrites already-proposed actions (\S\ref{subsec:wm-as-verifier}). Figure~\ref{fig:inference-pipeline} traces both modes side by side.

\input{figures/inference-pipeline-v3}

\subsection{World Model as Simulator}
\label{subsec:wm-as-simulator}

A natural use of a text world model at inference time is to simulate future states and use those predictions to guide action selection. The key design variable is the depth of simulation: shallow lookahead is cheap but myopic, while deep tree search explores exponentially more futures at correspondingly higher cost. Different methods strike this compute--quality balance in different ways.

\subsubsection{Shallow Lookahead}
\label{subsubsec:shallow-lookahead}

A first family of methods invokes the world model for a single step or a small number of steps before committing to an action, following a shared propose--simulate--score pattern: the agent proposes a set of candidates, the world model imagines the immediate consequence of each, and a scoring rule selects the best.

WMA~\citep{chae2025wma} instantiates this pattern by having the world model emit free-form natural-language descriptions of state differences, then scoring simulated outcomes for each candidate with a value function, attaining competitive performance against tree-search agents at substantially lower cost. WebDreamer~\citep{gu2025webdreamer} pursues the same one-step pattern in a training-free setting: it prompts the LLM to simulate the outcome of each candidate action and selects the best, but reports degradation as horizons extend, motivating smaller specialized world models. SimuRA~\citep{deng2025simuraworldmodeldrivensimulativereasoning} carries this pattern toward a more explicitly world-model-driven architecture: states are represented as natural-language summaries, and high-level simulated actions are scored over the resulting belief states before any action is executed, consistently improving task success over autoregressive planning on web browsing tasks.

A variant grounds LLM predictions in executable symbolic knowledge, combining flexible prediction with rule-based rigor~\citep{zhou2025walle2}: the system maintains a structured world state as symbolic rules (action rules, knowledge graphs, scene graphs), uses the LLM to predict preconditions and effects within this formalism, and applies Model Predictive Control for planning, attaining the highest reported success on ALFWorld.

\subsubsection{Deep Tree Search}
\label{subsubsec:tree-search}

When a single-step lookahead is insufficient, tree search provides a principled framework for multi-step planning, and text world models naturally serve as the transition function within such searches.

An early demonstration uses an LLM to generate candidate actions and estimate state transitions, while Monte Carlo Tree Search (MCTS) provides systematic exploration (LLM-MCTS;~\citealp{zhao2023llmmcts}); this separation of roles, with the LLM acting as world model and MCTS as search algorithm, established the pattern adopted by many subsequent works. RAP~\citep{hao2023rap} repurposes a single LLM as both world model and reasoning agent within MCTS, with rewards from action likelihood, state confidence, and self-evaluation; on Blocksworld, this enables a 33B model to outperform GPT-4 chain-of-thought, suggesting that structured search over an LLM world model can substitute for substantially larger models reasoning without search.

Incorporating environment feedback into the search loop extends this further: LATS~\citep{zhou2024lats} feeds self-reflection on failed trajectories back into MCTS, while Agent Q~\citep{putta2024agentq} closes a self-improving loop by combining tree search with offline DPO over both successful and failed trajectories, bridging inference-time and training-time use.
Pursuing the programmatic construction approach to its logical conclusion, Code World Models~\citep{lehrach2025codewm} translate game rules into executable Python code and run MCTS over this code-as-WM, matching or surpassing Gemini 2.5 Pro on the majority of evaluated games. TheoryCoder~\citep{ahmed2025theorycoder} extends this to bilevel planning: PDDL operators provide high-level abstract actions while LLM-synthesized Python functions implement low-level transitions, restricting search to the abstract space and only grounding transitions when needed.

Tree search can also be strengthened by improving the information supplied to the latent world model. LWM-Planner~\citep{holt2025improvingllmagentplanning} extracts task-critical atomic facts from interaction trajectories and uses them to augment action proposal, simulation, and value estimation in a recursive depth-limited search; grounding the search in an evolving fact memory lets the agent improve its planning online without weight updates.

\subsection{World Model as Verifier}
\label{subsec:wm-as-verifier}

Rather than driving action selection, a world model can act as a verifier on actions produced by the policy: the agent generates candidates through ordinary inference, and the world model predicts their consequences to decide whether each should be executed, replaced, or revised. This avoids expanding a search tree while mitigating the dynamics-blindness of LLMs~\citep{gupta2026worldworkflowsbenchmarkbringing}. Existing methods can be ordered by how strongly the verifier intervenes: from a single-action gate that accepts or rejects, to a ranker that selects among multiple candidates, to a corrector that triggers regeneration when no candidate is good enough.

\paragraph{Single-action gate}
The simplest instantiation is a single-action safety gate: fine-tuned LLM world models~\citep{li2025wordtoworld} simulate the outcome of each action and let the agent commit only when the prediction indicates success, improving web-task success and preventing catastrophic actions.

\paragraph{Ranking among candidates}
Most subsequent work exploits the verifier in a more discriminating mode, ranking among multiple candidates before any real execution. In software engineering, a Software World Reward model~\citep{sun2025sweworld} generates virtual test reports for candidate patches and selects the highest-scoring one before expensive real test execution. The same principle generalizes to GUI agents (CUWM;~\citealp{guan2026computerusingworldmodel}), where a two-stage world model predicts a textual description of UI changes and then realizes the next screenshot, letting the agent commit only to the candidate whose imagined outcome best matches the goal. In autonomous ML agent settings, where each training run costs hours, FOREAGENT~\citep{zheng2026foreagent} predicts which of two candidate solutions will perform better and uses confidence-gated pairwise prediction to physically execute only the winner, yielding substantial speedups on MLE-bench. For budget-constrained tool use, INTENT~\citep{liu2026intent} simulates ideal trajectories in which all tool calls succeed to extract the agent's latent plan and calibrate expected costs, allowing the agent to respect budget constraints while attaining high pass rates on cost-augmented StableToolBench.

\paragraph{Correction by regeneration}
When the entire candidate pool falls short, some methods replace selection with rewriting. WAC~\citep{shen2026wac} implements an iterate-until-confident loop: the world model simulates each candidate, a judge assigns confidence with rationale, and if all candidates fall below the threshold, the low-confidence actions and rationales are fed back to the action model for regeneration; on VisualWebArena, this consistently outperforms ranking-only baselines.

\begin{table}[t]
\centering
\small
\caption{Comparison of inference-time world model approaches. \textbf{Role}: Simulator = produces futures for action selection; Verifier = screens proposed actions. \textbf{Depth}: number of lookahead steps (1 = one-step, $k$ = multi-step, $\infty$ = full rollout, 0 = no rollout).}
\label{tab:inference-time-comparison}
\resizebox{0.8\columnwidth}{!}{%
\begin{tabular}{@{}llll@{}}
\toprule
\textbf{Method} & \textbf{Role} & \textbf{WM Type} & \textbf{Depth} \\
\midrule
WMA~\citep{chae2025wma} & Simulator (shallow) & Finetuned LLM & 1-step \\
WebDreamer~\citep{gu2025webdreamer} & Simulator (shallow) & Prompted LLM & 1-step \\
SimuRA~\citep{deng2025simuraworldmodeldrivensimulativereasoning} & Simulator (shallow) & Prompted LLM & $k$-step \\
WALL-E 2.0~\citep{zhou2025walle2} & Simulator (shallow) & LLM + symbolic rules & $k$-step \\
\midrule
LLM-MCTS~\citep{zhao2023llmmcts} & Simulator (search) & Prompted LLM & $k$-step \\
RAP~\citep{hao2023rap} & Simulator (search) & Prompted LLM & $k$-step \\
LATS~\citep{zhou2024lats} & Simulator (search) & Prompted LLM & $k$-step \\
Agent Q~\citep{putta2024agentq} & Simulator (search) & Prompted LLM & $k$-step \\
Code WM~\citep{lehrach2025codewm} & Simulator (search) & Executable code & $\infty$ \\
TheoryCoder~\citep{ahmed2025theorycoder} & Simulator (search) & Code (bilevel) & $k$-step \\
LWM-Planner~\citep{holt2025improvingllmagentplanning} & Simulator (search) & Prompted LLM + facts & $k$-step \\
\midrule
SWE-World~\citep{sun2025sweworld} & Verifier (rank) & Finetuned LLM & 1-step \\
CUWM~\citep{guan2026computerusingworldmodel} & Verifier (rank) & Two-stage GUI WM & 1-step \\
FOREAGENT~\citep{zheng2026foreagent} & Verifier (rank) & Implicit WM & 0 \\
INTENT~\citep{liu2026intent} & Verifier (rank) & Intent-aware & $\infty$ (ideal) \\
WAC~\citep{shen2026wac} & Verifier (rewrite) & Multi-agent & 1-step \\
\bottomrule
\end{tabular}%
}
\vspace{-10pt}
\end{table}

\subsection{Summary}
\label{subsec:inference-time-summary}

Table~\ref{tab:inference-time-comparison} consolidates the methods discussed in this section. A simulator is invoked when the policy cannot itself produce a usable candidate and external lookahead is needed, while a verifier presupposes a workable candidate set and only screens or rewrites it. What both share is a reliance on simulator fidelity; the simulator's predictions either drive action selection outright or back the verifier's judgments on candidate actions. The reach of either mode is therefore bounded by the same underlying question of how accurately the world model approximates the real environment, which we turn to next (\S\ref{sec:evaluation}).

%% file: figures/inference-pipeline-v3.tex
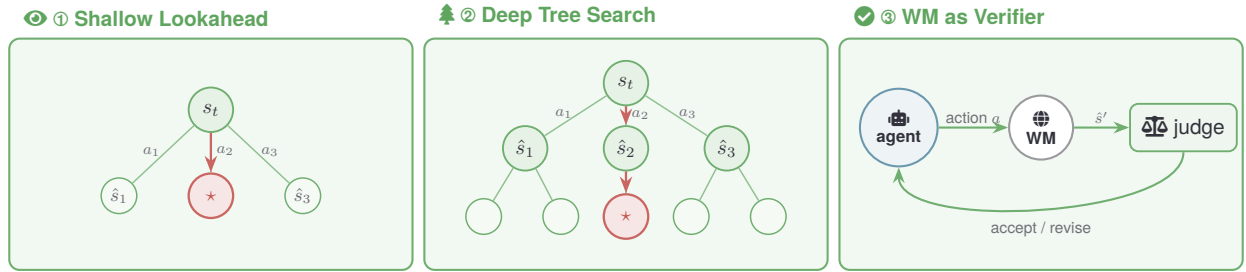
\begin{figure}
    \centering
    \resizebox{0.99\textwidth}{!}{%
    \begin{tikzpicture}[
        >=Stealth, font=\sffamily,
        panel/.style={rectangle, rounded corners=5pt,
            draw=morInfer!90, fill=morInfer!8, line width=1pt,
            minimum width=6.8cm, minimum height=3.9cm},
        ptitle/.style={font=\sffamily\normalsize\bfseries, text=morInfer, anchor=south west},
        sn/.style={circle, draw=morInfer!90, fill=morInfer!15, line width=0.9pt,
            minimum size=0.75cm, font=\sffamily\fontsize{10}{11}\selectfont, text=morInk, inner sep=0pt},
        leaf/.style={circle, draw=morInfer!90, fill=morInfer!6, line width=0.7pt,
            minimum size=0.6cm, font=\sffamily\fontsize{9}{10}\selectfont, text=morInk!75, inner sep=0pt},
        best/.style={circle, draw=morEval!90, fill=morEval!15, line width=1.1pt,
            minimum size=0.75cm, font=\sffamily\fontsize{10}{11}\selectfont\bfseries, text=morEval, inner sep=0pt},
        edge/.style={-, line width=0.7pt, color=morInfer!70},
        bedge/.style={->, line width=1.1pt, color=morEval!90},
        wm/.style={circle, draw=morInk!50, fill=white, line width=1pt,
            minimum size=0.95cm, font=\sffamily\fontsize{8}{9}\selectfont\bfseries, text=morInk, align=center},
        ag/.style={circle, draw=morAgent!90, fill=morAgent!10, line width=1pt,
            minimum size=0.95cm, font=\sffamily\fontsize{8}{9}\selectfont\bfseries, text=morInk, align=center},
        store/.style={cylinder, shape border rotate=90, aspect=0.25,
            draw=morInk!50, fill=morInfer!6, line width=0.9pt,
            minimum width=1.4cm, minimum height=1.0cm,
            font=\sffamily\fontsize{10}{11}\selectfont, text=morInk, align=center},
        verif/.style={rectangle, rounded corners=4pt, draw=morInfer!90, fill=morInfer!10,
            minimum width=1.8cm, minimum height=0.8cm,
            font=\sffamily\fontsize{10}{11}\selectfont, text=morInk, align=center, line width=0.9pt},
        arr/.style={->, line width=1.0pt, color=morInfer!90},
        lbl/.style={font=\sffamily\fontsize{7.5}{8.5}\selectfont, text=morInk!70, inner sep=1.5pt},
    ]
    \node[panel] (P1) at (-7.0, 0) {};
    \node[ptitle] at ([xshift=4pt, yshift=2pt]P1.north west) {\faEye~\textcircled{\scriptsize 1}~Shallow Lookahead};
    \node[sn]   (S-R)  at ($(P1.center)+( 0.0, 0.75)$) {$s_t$};
    \node[leaf] (S-L1) at ($(P1.center)+(-1.55, -0.70)$) {$\hat{s}_1$};
    \node[best] (S-L2) at ($(P1.center)+( 0.0, -0.70)$) {$\star$};
    \node[leaf] (S-L3) at ($(P1.center)+( 1.55, -0.70)$) {$\hat{s}_3$};
    \draw[edge] (S-R) -- node[lbl, pos=0.45, left=1pt] {$a_1$} (S-L1);
    \draw[edge] (S-R) -- node[lbl, pos=0.45, right=1pt] {$a_2$} (S-L2);
    \draw[edge] (S-R) -- node[lbl, pos=0.45, right=1pt] {$a_3$} (S-L3);
    \draw[bedge] (S-R) -- (S-L2);

    \node[panel] (P2) at (0, 0) {};
    \node[ptitle] at ([xshift=4pt, yshift=2pt]P2.north west) {\faTree~\textcircled{\scriptsize 2}~Deep Tree Search};
    \node[sn] (R)  at ($(P2.center)+( 0.0, 1.20)$) {$s_t$};
    \node[sn] (M1) at ($(P2.center)+(-1.7, 0.10)$) {$\hat{s}_1$};
    \node[sn] (M2) at ($(P2.center)+( 0.0, 0.10)$) {$\hat{s}_2$};
    \node[sn] (M3) at ($(P2.center)+( 1.7, 0.10)$) {$\hat{s}_3$};
    \draw[edge] (R) -- node[lbl, pos=0.45, left=1pt]  {$a_1$} (M1);
    \draw[edge] (R) -- node[lbl, pos=0.45, right=1pt] {$a_2$} (M2);
    \draw[edge] (R) -- node[lbl, pos=0.45, right=1pt] {$a_3$} (M3);
    \node[leaf] (L1) at ($(P2.center)+(-2.40, -1.05)$) {};
    \node[leaf] (L2) at ($(P2.center)+(-1.10, -1.05)$) {};
    \node[best] (L3) at ($(P2.center)+( 0.00, -1.05)$) {$\star$};
    \node[leaf] (L4) at ($(P2.center)+( 1.10, -1.05)$) {};
    \node[leaf] (L5) at ($(P2.center)+( 2.40, -1.05)$) {};
    \draw[edge] (M1) -- (L1); \draw[edge] (M1) -- (L2);
    \draw[edge] (M2) -- (L3);
    \draw[edge] (M3) -- (L4); \draw[edge] (M3) -- (L5);
    \draw[bedge] (R) -- (M2); \draw[bedge] (M2) -- (L3);

    \node[panel] (P3) at (7.0, 0) {};
    \node[ptitle] at ([xshift=4pt, yshift=2pt]P3.north west) {\faCheckCircle~\textcircled{\scriptsize 3}~WM as Verifier};
    \node[ag]    (V-ag) at ($(P3.center)+(-2.4, 0.45)$) {\faRobot\\agent};
    \node[wm]    (V-wm) at ($(P3.center)+( 0.0, 0.45)$) {\faGlobe\\WM};
    \node[verif] (V-vf) at ($(P3.center)+( 2.4, 0.45)$) {\faBalanceScale~judge};
    \draw[arr] (V-ag) -- node[lbl, above] {action $a$} (V-wm);
    \draw[arr] (V-wm) -- node[lbl, above] {$\hat{s}'$} (V-vf);
    \draw[arr] (V-vf.south) .. controls +(0,-1.1) and +(0,-1.1) ..
        node[lbl, below, yshift=-4pt] {accept / revise} (V-ag.south);
    \end{tikzpicture}}
    \caption{Inference-time roles of a text world model. \textbf{WM as simulator} (left, centre): \textbf{shallow lookahead} imagines the immediate consequence of each candidate action and picks the best; \textbf{deep tree search} uses the WM as transition function for multi-step rollouts. \textbf{WM as verifier} (right): the WM predicts the consequence of a proposed action, and a judge accepts or sends it back for revision.}
    \label{fig:inference-pipeline}
\end{figure}

%% file: sections/06-evaluation.tex
\section{Evaluation}
\label{sec:evaluation}

The preceding sections have established that text world models can be built (\S\ref{sec:building}), used to train agents (\S\ref{sec:training-time}), and deployed at inference time (\S\ref{sec:inference-time}). This section addresses how they are evaluated, as illustrated in Figure~\ref{fig:evaluation-pipeline}. Two perspectives emerge from a role inversion of the world model itself: it can be the object of evaluation, with metrics measuring prediction accuracy, consistency, and task utility (\S\ref{subsec:eval-wm}); or it can be the evaluation tool, with simulated users and environments serving as benchmarking substrates for agents (\S\ref{subsec:wm-as-eval}).

\subsection{Evaluating World Models Themselves}
\label{subsec:eval-wm}

When the world model itself is the object of evaluation, two complementary questions arise: how accurately does it predict the next state (\S\ref{subsubsec:prediction-accuracy}), and does that accuracy translate into downstream task utility (\S\ref{subsubsec:task-driven-eval})?

\subsubsection{Prediction Accuracy and Consistency}
\label{subsubsec:prediction-accuracy}

The most direct way to evaluate a world model is to examine whether it can accurately predict subsequent states. Existing metrics extend along two axes: prediction horizon, from single-step fidelity to multi-step coherence under rollout, and observation regime, from purely textual settings to multimodal and partially observable ones.

\paragraph{Single-step metrics}
The primary single-step metric is exact-match (EM) accuracy, which checks whether each predicted state attribute matches the ground-truth post-action state. ByteSized32-State-Prediction~\citep{wang2024canlms} introduced this evaluation over a large set of state transitions from text games, and finds that even strong frontier models attain only modest EM on non-trivial transitions, with environment-driven changes and arithmetic-heavy tasks proving particularly difficult, establishing that prompted LLMs are unreliable world simulators. Supervised fine-tuning~\citep{li2025wordtoworld} closes this gap substantially: smaller open-source models achieve near-saturated EM on structured environments (ALFWorld, SciWorld) and markedly lower performance on open-ended ones (WebShop). The sizeable gap between prompted and fine-tuned world models is one of the most consistent empirical findings in this field.

\paragraph{Multi-step consistency}
Single-step accuracy does not guarantee long-horizon reliability, as errors compound over sequential predictions. The consistency ratio CR~\citep{li2025wordtoworld} measures what fraction of trajectories that succeed in the world model also succeed in the real environment; it remains high in structured environments but drops in open-ended settings without anchoring techniques. Probing experiments on computer-use agents~\citep{mei2025rwom} further quantify this degradation: single-step next-state identification is comparatively reliable, but full-procedure planning alignment lags well behind. $CR_{\text{pw}}$~\citep{huang2026stateconsistencybehaviorconsistency} tightens this measurement at the per-trajectory level, and motivates a behavior-consistency training signal for cases where text-level similarity fails to reflect decision preservation.

\paragraph{Multimodal and partially observable settings}
Recent benchmarks revisit the same fidelity questions in vision-language and embodied regimes where observations are often partial. WorldPrediction~\citep{chen2025worldprediction} reformulates evaluation as discriminative tasks: given initial and final states, models must select the correct action from counterfactual distractors, with even the strongest models scoring well below human performance. ENACT~\citep{wang2025enactevaluatingembodiedcognition} confirms this gap in embodied settings, where frontier models perform near chance on multi-step forward reasoning while humans remain highly accurate.

\input{figures/evaluation-pipeline-v2}

\subsubsection{Task-Driven Evaluation}
\label{subsubsec:task-driven-eval}

Beyond raw prediction accuracy, several works evaluate whether world models are useful for downstream tasks, since strong next-state scores need not translate into agent utility. Existing approaches differ in what notion of usefulness they probe: capability across multiple downstream uses, executability of a symbolic specification, or whether task success even reflects world understanding at all. \citet{yang2026evalwm} propose a three-task framework spanning policy verification, action generation, and long-horizon planning, and find that performance degrades substantially with horizon and that capabilities are inconsistent across tasks: models strong on verification are not necessarily strong on planning. Treating world models as compilers of executable symbolic theories, Text2World~\citep{hu2025text2world} tests whether LLMs can generate valid PDDL domain models from natural language descriptions, scoring both executability and component-level F1 over predicates, parameters, preconditions, and effects, and finds that even the strongest models score modestly on precondition and effect F1 without iterative error correction. Decoupling task completion from environment understanding, Task2Quiz~\citep{liu2026task2quiz} introduces task success rate (TSR) and a separate environment understanding score (EUS) measured via trajectory-conditioned QA. TSR drops sharply with task difficulty while EUS remains comparatively stable, demonstrating that an agent may complete tasks without truly understanding the world. This dissociation challenges the common assumption that task performance is a sufficient proxy for world model quality.

\begin{takeaway}
Prediction accuracy and task utility are not interchangeable: single-step EM saturates on structured environments but collapses under multi-step rollout, and even when prediction accuracy is high, task success and environment understanding can dissociate (Task2Quiz), so any single metric overstates competence in some regime. The shared limitation across this subsection is that fine-grained metrics (CR, $CR_{\text{pw}}$, executability F1, EUS) require either ground-truth rollouts or carefully constructed probes, neither of which scales easily to new domains; this limits cross-domain comparison and motivates the simulator-based evaluation paradigm of \S\ref{subsec:wm-as-eval}.
\end{takeaway}

\subsection{World Models as Evaluation Environments}
\label{subsec:wm-as-eval}

A complementary use of text world models is as evaluation tools: simulating users, environments, or entire interaction scenarios to benchmark agent capabilities. Increasingly, however, the quality of the simulator itself also becomes a first-class evaluation target: if simulated users are too cooperative or simulated environments are insufficiently faithful, downstream benchmark results can be misleading. We therefore organize this subsection along two lines: benchmark design, which asks what to evaluate and how, and simulator validity, which asks whether the simulator is a faithful proxy for reality.

\subsubsection{Benchmark Design}
\label{subsubsec:benchmark-design}

Existing benchmarks differ in what part of the world the simulator covers: the system environment (interfaces, programs), the human user, or domain-specific verticals where state ties to non-generic signals such as sensors or organizational telemetry.

\paragraph{Environment simulation benchmarks}
Environment-centric benchmarks evaluate how faithfully models track interface- or program-level state evolution, with two common foci of semantic fidelity and cross-environment transfer. MobileWorldBench~\citep{li2025mobileworldbench} redefines GUI world modeling from pixel-level next-frame prediction to semantic-level natural language prediction, and shows that a fine-tuned 8B model used as a semantic world model boosts AndroidWorld task success by 7.4\%. AutoEnv~\citep{zhang2025autoenv} instead generates programmatic environments for cross-environment transfer evaluation, testing whether agents trained in one set of environments generalize to unseen ones.

\paragraph{User simulation benchmarks}
Another body of work foregrounds simulated users, with natural-language turns carrying intent and procedural constraints while tool-mediated backends supply latent state. Existing work moves from static single-actor simulation toward dual-actor and long-horizon settings. $\tau$-bench~\citep{yao2024taubenchbenchmarktoolagentuserinteraction} introduces tool--agent--user interaction with end-state verification in a hidden backend, and $\tau^2$-Bench~\citep{barres2025tau2bench} extends it to a dual-control Dec-POMDP in which both the agent and a simulated user can act. FUSE~\citep{kudrinskii2025fuse} generalizes this into a closed-loop user--agent--environment simulator with configurable user and environment archetypes, and additionally evaluates Procedure Alignment and simulation faithfulness. LifeSim~\citep{duan2026lifesimlonghorizonuserlife} extends user simulation toward long-horizon personal dynamics with a Belief-Desire-Intention user model spanning 8 life domains, enabling evaluation of implicit intent recognition and long-term preference tracking. Treating emotion as part of the simulated user, LEWM~\citep{song2025largeemotionalworldmodel} jointly predicts the next environment state and the next emotional state, and shows that removing emotional context degrades subjective-task accuracy by up to 8\% while affecting objective tasks by only $\sim$1\%.

\paragraph{Domain-specific benchmarks}
A final group anchors evaluation in vertical settings where world state ties to non-generic signals such as instruments, geospatial records, or organizational telemetry rather than dialogue or GUI abstractions. \citet{li2025disasterwm} apply text world models as ``virtual sensors'' that fuse seismic, geospatial, and street-view signals to predict human-perceived earthquake intensity. \citet{gupta2026worldworkflowsbenchmarkbringing} apply language models as enterprise-level world models, and find that current LLMs often fail to predict latent cascading side effects in partially observable organizational systems.

\subsubsection{Simulator Validity}
\label{subsubsec:simulator-validity}

As simulated users and environments become central to benchmarking, a natural follow-up question is whether the simulator itself is a faithful proxy for reality. We organize existing work along two angles: faithfulness to real users, and the structural properties (efficiency, feedback richness) that determine how informative an interaction-based evaluation can be.

\paragraph{Faithfulness to real users}
Instead of prompting an assistant model to ``act like a user,'' \citet{naous2025flipping} trains dedicated User Language Models on real human--assistant conversations and shows that more realistic user simulation substantially lowers apparent assistant performance, exposing weaknesses hidden by overly cooperative simulators. SimulatorArena~\citep{dou2025simulatorarenausersimulatorsreliable} directly tests whether simulated users are reliable substitutes for human evaluation by measuring alignment between simulator-based and human ratings. \citet{zhou2026mindsim2realgapuser} compare 31 LLM simulators against 451 real participants over 165 tasks, introduce a User-Sim Index, and find that many simulators create an ``easy mode'' by being excessively polite and forgiving. \citet{seshadri2026lostsimulationllmsimulatedusers} reach a similarly cautionary conclusion in $\tau$-Bench retail tasks, additionally finding that simulators are unevenly faithful across demographic and dialectal groups, making simulator validity a fairness question.

\paragraph{Interaction structure and cost}
Once evaluation is framed as an interaction, it must also account for the efficiency and richness of that process. IDRBench~\citep{feng2026idrbench} jointly measures interaction benefit (report quality improvement) and cost (query turns and tokens), and finds that early planning-stage clarification yields the best benefit-to-cost ratio. \citet{yue2026interactivebench} formalize LLM assessment as budget-constrained sequential decision-making and show that interactive evaluation reveals capabilities invisible to static benchmarks, with pass@$k$ substantially underestimating ability on tasks requiring iterative probing. RECODE-H~\citep{miao2025recodeh} introduces five hierarchical levels of human feedback (L0--L4) and finds that richer feedback significantly improves GPT-5's performance, indicating that the granularity of feedback supplied by the simulator materially shapes what the benchmark measures.

\subsection{Summary: The Evaluation Landscape}
\label{subsec:eval-summary}

This section has covered two complementary directions. The first examines the world model as the object of evaluation, since both training-time and inference-time uses (\S\ref{sec:training-time}, \S\ref{sec:inference-time}) ultimately inherit its prediction quality: a simulator that drifts during rollout can mislead the agent, while a verifier judging from an incorrectly predicted state may reject the correct action. Accordingly, existing metrics have moved beyond surface-level token matching toward downstream utility, with consistency ratios, behavior preservation, and decoupled understanding scores increasingly replacing exact-match accuracy. This shift parallels the reward designs discussed in \S\ref{sec:rl-wm}, where supervision likewise moves from token fidelity toward behavior-level signals. Together, these trends suggest that ``how to score a world model'' and ``how to train one'' are converging along the same axis.

In parallel, the world model has itself become an evaluation protocol, extending assessment beyond static input--output checks to settings that were previously difficult to evaluate: dynamic multi-turn interaction, dual-actor decision-making, simulated users, and domain-specific verticals where real-world evaluation is impractical. The cost of this expansion is that benchmark conclusions are only as reliable as the simulator that produces them. Recent work shows that this gap is non-trivial: prompted assistant LLMs can systematically overestimate agent competence by being overly cooperative and stylistically uniform~\citep{naous2025flipping, zhou2026mindsim2realgapuser}, and the gap varies across demographic and dialectal groups. Simulator faithfulness has therefore become a first-class evaluation target alongside the agent it is meant to score.

%% file: figures/evaluation-pipeline-v2.tex
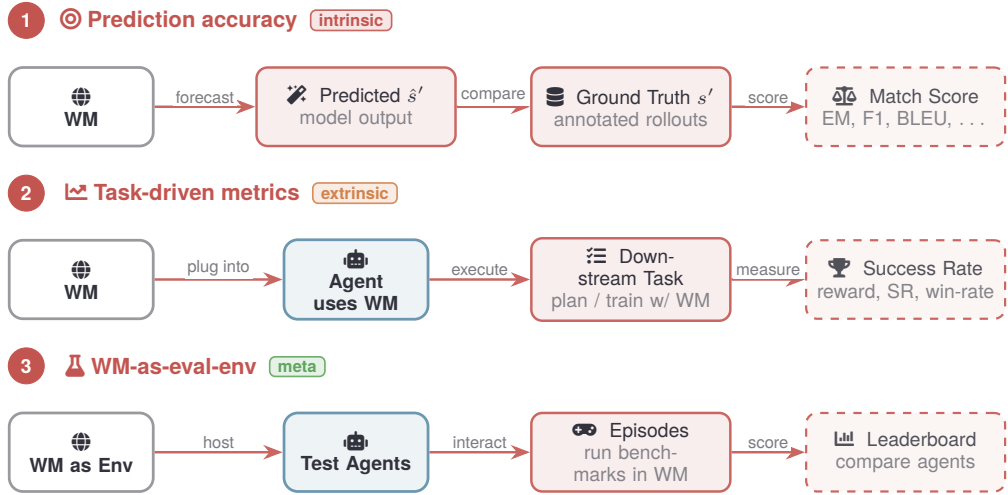
\begin{figure}
    \centering
    \resizebox{0.8\textwidth}{!}{%
    \begin{tikzpicture}[
        >=Stealth, font=\sffamily,
        numcircle/.style={circle, draw=morEval, fill=morEval, text=white,
            font=\sffamily\bfseries\fontsize{7}{8}\selectfont, inner sep=0pt, minimum size=13pt},
        rowtitle/.style={font=\sffamily\footnotesize\bfseries, text=morEval, anchor=west},
        block/.style={rectangle, rounded corners=4pt,
            minimum width=2.4cm, minimum height=1.0cm, text width=2.3cm, align=center,
            font=\sffamily\scriptsize, text=morInk,
            draw=morEval!90, fill=morEval!10, line width=0.9pt},
        agentblock/.style={rectangle, rounded corners=5pt,
            minimum width=1.7cm, minimum height=1.0cm, text width=1.6cm, align=center,
            font=\sffamily\scriptsize\bfseries, text=morInk,
            draw=morAgent!90, fill=morAgent!10, line width=1pt},
        wmblock/.style={rectangle, rounded corners=5pt,
            minimum width=1.7cm, minimum height=1.0cm, text width=1.6cm, align=center,
            font=\sffamily\scriptsize\bfseries, text=morInk,
            draw=morInk!50, fill=white, line width=1pt},
        outblock/.style={rectangle, rounded corners=4pt,
            minimum width=2.4cm, minimum height=1.0cm, text width=2.3cm, align=center,
            font=\sffamily\scriptsize, text=morInk!85,
            draw=morEval!90, fill=morEval!6, line width=0.8pt, dashed},
        tag/.style={rectangle, rounded corners=2pt, draw=#1!90, fill=#1!15,
            font=\sffamily\fontsize{6}{7}\selectfont\bfseries, text=#1,
            inner xsep=3pt, inner ysep=1.5pt, line width=0.6pt},
        arr/.style={->, line width=0.9pt, color=morEval!90},
        lbl/.style={font=\sffamily\fontsize{6}{7}\selectfont, text=morInk!65, fill=white, inner sep=1pt},
    ]
    \node[numcircle] (n1) at (-0.7, 0) {1};
    \node[rowtitle, right=2pt of n1] {\faBullseye~Prediction accuracy~~\tikz[baseline=-0.6ex]\node[tag=morEval] {intrinsic};};
    \node[wmblock]  (e1) at (0,    -1.1) {\faGlobe\\WM};
    \node[block]    (e2) at (3.5,  -1.1) {\faMagic~~Predicted $\hat{s}'$\\\textcolor{morInk!60}{model output}};
    \node[block]    (e3) at (7.0,  -1.1) {\faDatabase~~Ground Truth $s'$\\\textcolor{morInk!60}{annotated rollouts}};
    \node[outblock] (e4) at (10.5, -1.1) {\faBalanceScale~~Match Score\\\textcolor{morInk!60}{EM, F1, BLEU, $\dots$}};
    \draw[arr] (e1) -- node[lbl, above] {forecast} (e2);
    \draw[arr] (e2) -- node[lbl, above] {compare} (e3);
    \draw[arr] (e3) -- node[lbl, above] {score} (e4);

    \node[numcircle] (n2) at (-0.7, -2.2) {2};
    \node[rowtitle, right=4pt of n2] {\faChartLine~Task-driven metrics~~\tikz[baseline=-0.6ex]\node[tag=morTrain] {extrinsic};};
    \node[wmblock]    (t1) at (0,    -3.3) {\faGlobe\\WM};
    \node[agentblock] (t2) at (3.5,  -3.3) {\faRobot\\Agent uses WM};
    \node[block]      (t3) at (7.0,  -3.3) {\faTasks~~Downstream Task\\\textcolor{morInk!60}{plan / train w/ WM}};
    \node[outblock]   (t4) at (10.5, -3.3) {\faTrophy~~Success Rate\\\textcolor{morInk!60}{reward, SR, win-rate}};
    \draw[arr] (t1) -- node[lbl, above] {plug into} (t2);
    \draw[arr] (t2) -- node[lbl, above] {execute} (t3);
    \draw[arr] (t3) -- node[lbl, above] {measure} (t4);

    \node[numcircle] (n3) at (-0.7, -4.4) {3};
    \node[rowtitle, right=4pt of n3] {\faFlask~WM-as-eval-env~~\tikz[baseline=-0.6ex]\node[tag=morInfer] {meta};};
    \node[wmblock]    (v1) at (0,    -5.5) {\faGlobe\\WM as Env};
    \node[agentblock] (v2) at (3.5,  -5.5) {\faRobot\\Test Agents};
    \node[block]      (v3) at (7.0,  -5.5) {\faGamepad~~Episodes\\\textcolor{morInk!60}{run benchmarks in WM}};
    \node[outblock]   (v4) at (10.5, -5.5) {\faChartBar~~Leaderboard\\\textcolor{morInk!60}{compare agents}};
    \draw[arr] (v1) -- node[lbl, above] {host} (v2);
    \draw[arr] (v2) -- node[lbl, above] {interact} (v3);
    \draw[arr] (v3) -- node[lbl, above] {score} (v4);

    \end{tikzpicture}}
    \caption{Three evaluation paradigms for text world models: intrinsic prediction accuracy against ground-truth next states, extrinsic task-driven metrics where the world model is plugged into a downstream agent, and meta evaluation where the world model itself serves as a benchmarking environment for agents.}
    \label{fig:evaluation-pipeline}
\end{figure}

%% file: sections/07-open-problems.tex
\section{Open Problems and Future Directions}
\label{sec:open-problems}

\subsection{World Model--Policy Coupling}
\label{subsec:coupling}

A choice that cuts across construction, training, and inference, but is rarely articulated as a choice, is how tightly the world model is bound to the policy that uses it. Three regimes recur: a single LLM playing both roles via prompting~\citep{zhang2025earlyexperience}; a shared backbone with role-specific adapters or prompts; and fully decoupled models, where a dedicated transition predictor~\citep{chae2025wma, chen2025dreamgym} is queried by an arbitrary downstream agent. Sharing parameters eliminates state-language drift and deployment overhead, but world-model errors enter the policy gradient directly and the two objectives compete for capacity, often invisibly. Decoupling reverses both effects: components scale independently and one world model can serve many policies, but the policy must consume states in the exact form the world model emits, and any vocabulary or granularity drift silently degrades rollouts. The middle regime preserves a shared state space while allowing separate optimization signals, at the cost of balancing two losses.

There is no a priori winner: when the agent is the only consumer, full sharing is parsimonious; when the world model is itself a research artifact (queryable, benchmark-able, reusable across policies), decoupling pays off. A practical implication is that papers reporting world-model accuracy in isolation and papers reporting downstream agent reward are not measuring the same object, even on overlapping benchmarks.

\subsection{Reasoning World Models}
\label{subsec:reasoning-wm}

Most text world models perform \emph{direct} next-state prediction, mapping a state--action pair to a successor without intermediate reasoning. This suffices for simple dynamics but breaks down when prediction itself requires multi-step inference: predicting code execution requires tracing control flow, scientific simulation requires causal reasoning about physical laws, and user modeling requires inferring latent intent. In such settings, world modeling is fundamentally a reasoning task. Recent work already shows reasoning models serving as effective world simulators~\citep{li2025simia, yu2025dynathink}, and CoT objectives consistently improve transition prediction~\citep{sun2025sweworld, chen2025dreamgym}, pointing to a clear direction: \emph{world models should be equipped with explicit reasoning capabilities}.

Standard trajectory-level SFT encourages surface pattern matching rather than reasoning. Three underexplored paradigms exist: (1)~\emph{CoT distillation} from stronger teachers~\citep{sun2025sweworld}; (2)~\emph{joint reasoning--prediction objectives} combining CoT supervision with transition loss~\citep{chen2025dreamgym, yu2025dynamind}; and (3)~\emph{RL with process rewards}, which could borrow from agentic RL techniques that elicit deliberate or meta-reasoning behaviors~\citep{shang2025rstar2agent, zhang2025rlvmr}. Adapting such signals to reward predictions achieved \emph{via} reasoning rather than memorization, along with characterizing how reasoning depth should scale with prediction difficulty, remains largely unexplored.

\subsection{Architecture and Integration}
\label{subsec:arch-integration}

\paragraph{Unified cross-lifecycle architectures}
\label{subsec:unified-arch}
Current text world models are typically designed for a single lifecycle stage: construction (\S\ref{sec:building}), training (\S\ref{sec:training-time}), or inference (\S\ref{sec:inference-time}).
A world model trained for next-state prediction may not be directly useful for tree search, and one optimized for training rollouts may not serve as an effective evaluator.
Future work should explore \emph{unified architectures} that serve multiple lifecycle stages with a single model, potentially through multi-task training objectives that simultaneously optimize prediction accuracy, planning utility, and evaluation capability.

\paragraph{World-model-aware agent design}
\label{subsec:wm-aware-arch}
Current agents fail to leverage world models~\citep{qian2025failleverage}, with some invoking them in fewer than 1\% of episodes, suggesting that agent architectures are not designed to effectively \emph{integrate} world model predictions.
Future agent architectures should be explicitly designed around world model capabilities, with mechanisms for:
(1) deciding \emph{when} to query the world model (building on AVIC's~\citep{yu2025avic} adaptive gating),
(2) \emph{how much} to trust its predictions (building on FOREAGENT's~\citep{zheng2026foreagent} confidence calibration), and
(3) \emph{how to correct} when predictions are wrong (building on WAC's~\citep{shen2026wac} closed-loop correction).

\subsection{Grounding, Adaptation, and Generalization}
\label{subsec:grounding-adaptation}

\paragraph{Grounded text world models}
\label{subsec:grounded-wm}
Most text world models operate in \emph{digital} environments.
Connecting these to physical reality, by grounding text predictions in sensor data, physical constraints, and real-world consequences, remains a fundamental challenge.
The disaster assessment work~\citep{li2025disasterwm} provides an early example, but systematic approaches to grounding text world models in physical observations are largely unexplored.
Multimodal world models like AVIC~\citep{yu2025avic} and MobileWorldBench~\citep{li2025mobileworldbench} represent steps toward bridging this gap.

\paragraph{Continual learning and adaptation}
\label{subsec:continual}
Real-world environments are non-stationary: websites update, APIs change, and user preferences drift.
While test-time adaptations~\citep{chen2025testtimeadapt, wei2025evomemory} address short-term adaptation, long-term continual learning of text world models, which must maintain accuracy on old environments while adapting to new ones, remains unexplored.
PAHF's~\citep{liang2026pahf} dual-channel feedback loop provides theoretical insights on adaptation rates, but practical continual learning systems for text world models have yet to be developed.

%% file: sections/08-conclusion.tex
\section{Conclusion}
\label{sec:conclusion}

In this survey, we presented the first systematic review of text world models for LLM-based agents, organizing the field along both a formal two-axis framework, spanning state representation and grounding domain, and the full agent lifecycle, from construction through training-time and inference-time application to evaluation. We characterized how LLM-as-WM and Code-as-WM approaches instantiate the transition function under different assumptions about data, fidelity, and verifiability, and how the resulting models support agents at training and inference time. By offering a unified perspective across these dimensions and surfacing the corresponding open challenges, we hope this survey can serve as a valuable resource for advancing research in this rapidly evolving area.